%% file: main.tex
\documentclass[10pt,journal,compsoc]{IEEEtran}
\ifCLASSOPTIONcompsoc
  \usepackage[nocompress]{cite}
\else
  \usepackage{cite}
\fi

\usepackage[normalem]{ulem}

\usepackage[table]{xcolor}
\usepackage{mathtools}
\usepackage{booktabs}
\usepackage{overpic}
\usepackage{paralist}
\usepackage{multirow}
\usepackage{tikz}
\usepackage[colorlinks]{hyperref}
\usepackage[accsupp]{axessibility}  % Improves PDF readability for those with disabilities.
\usepackage{algorithm, algorithmic}
\algsetup{linenosize=\small}
\usepackage{tablefootnote}
\usepackage{tabularx}
\usepackage{graphicx}
\usepackage[flushleft]{threeparttable} % for tablenotes
\DeclarePairedDelimiter\ceil{\lceil}{\rceil}

\newcommand{\reffig}[1]{Fig.~\ref{#1}}
\newcommand{\reftab}[1]{Tab.~\ref{#1}}
\newcommand{\refsec}[1]{Sec.~\ref{#1}}
\newcommand*{\Scale}[2][4]{\scalebox{#1}{$#2$}}%

\newcommand{\specialcell}[2][c]{%
  \begin{tabular}[#1]{@{}c@{}}#2\end{tabular}}
\ifCLASSINFOpdf
\else
\fi
\usepackage{amsmath}
\usepackage{array}
\ifCLASSOPTIONcompsoc
 \usepackage[caption=false,font=normalsize,labelfont=sf,textfont=sf]{subfig}
\else
 \usepackage[caption=false,font=footnotesize]{subfig}
\fi
\usepackage{url}
\begin{document}
\title{Audio2Gestures: Generating Diverse Gestures from Audio}

\author{Jing~Li,
        Di~Kang,
        Wenjie~Pei,
        Xuefei~Zhe,
        Ying~Zhang,
        Linchao~Bao,
        and~Zhenyu~He
\thanks{J. Li, W. Pei and Z. He are with Harbin Institute of Technology, Shenzhen.}% <-this % stops a space
\thanks{D. Kang, X. Zhe, Y. Zhang and L. Bao are with Tencent AI Lab, Shenzhen.}
\thanks{L. Bao and Z. He are the corresponding authors.}
}

% and the end of the author line. \textit{i.e.}, if you had this:

\markboth{Journal of \LaTeX\ Class Files,~Vol.~14, No.~8, August~2015}%
{Shell \MakeLowercase{\textit{et al.}}: Bare Demo of IEEEtran.cls for IEEE Journals}



\maketitle

\begin{abstract}
People may perform diverse gestures affected by various mental and physical factors when speaking the same sentences.
This inherent one-to-many relationship makes co-speech gesture generation from audio particularly challenging.
Conventional CNNs/RNNs assume one-to-one mapping, and thus tend to predict the average of all possible target motions, easily resulting in plain/boring motions during inference. 
So we propose to explicitly model the one-to-many audio-to-motion mapping by splitting the cross-modal latent code into shared code and motion-specific code.
The shared code is expected to be responsible for the motion component that is more correlated to the audio 
while the motion-specific code is expected to capture diverse motion information that is more independent of the audio.
However, splitting the latent code into two parts poses extra training difficulties. 
Several crucial training losses/strategies, including relaxed motion loss, bicycle constraint, and diversity loss, are designed to better train the VAE.

% Thorough experiments have been conducted to 1) verify the effectiveness and universality of the proposed split latent space and to 2) better evaluate the quality of the generated motions with various motion metrics.
Experiments on both 3D and 2D motion datasets verify that our method generates more realistic and diverse motions than previous state-of-the-art methods, quantitatively and qualitatively.
Besides, our formulation is compatible with discrete cosine transformation (DCT) modeling and other popular backbones (\textit{i.e.} RNN, Transformer).
As for motion losses and quantitative motion evaluation, we find structured losses/metrics (\textit{e.g.} STFT) that consider temporal and/or spatial context complement the most commonly used point-wise losses (\textit{e.g.} PCK), resulting in better motion dynamics and more nuanced motion details.
Finally, we demonstrate that our method can be readily used to generate motion sequences with user-specified motion clips on the timeline.
% Code and more results are at \url{https://jingli513.github.io/audio2gestures}.
\end{abstract}


% \fi
\IEEEpeerreviewmaketitle

\section{Introduction}

% \dknote{check table ref format}


% \IEEEPARstart{I}{n} the real world, co-speech gestures help express oneself better, and in the virtual world, it makes a talking avatar act more vividly.
\IEEEPARstart{G}{}enerating vivid human-like co-speech gestures is of great importance for producing attractive avatars that people are willing to interact with.
There has been a surging demand for generating realistic human motions for given audio clips recently. 
However, this problem is very challenging because of the complicated one-to-many relationship between audio and motion.
For example, a speaker may perform different gestures under different conditions (\textit{e.g.} happy/peaceful mood, standing/sitting state, or different environments)
when speaking the same words due to different mental and physical states.
\par
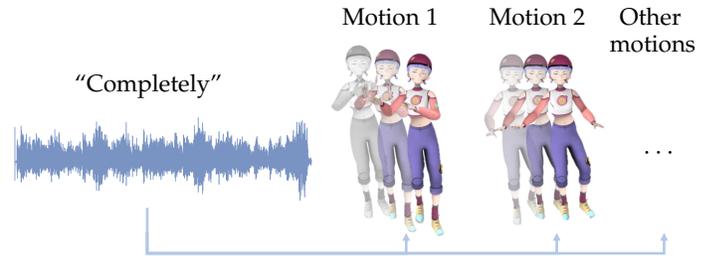
\begin{figure}[!t]
\centering
\begin{overpic}[trim=2.7cm 6.3cm 2cm 5.5cm,clip,width=1\linewidth,grid=false]{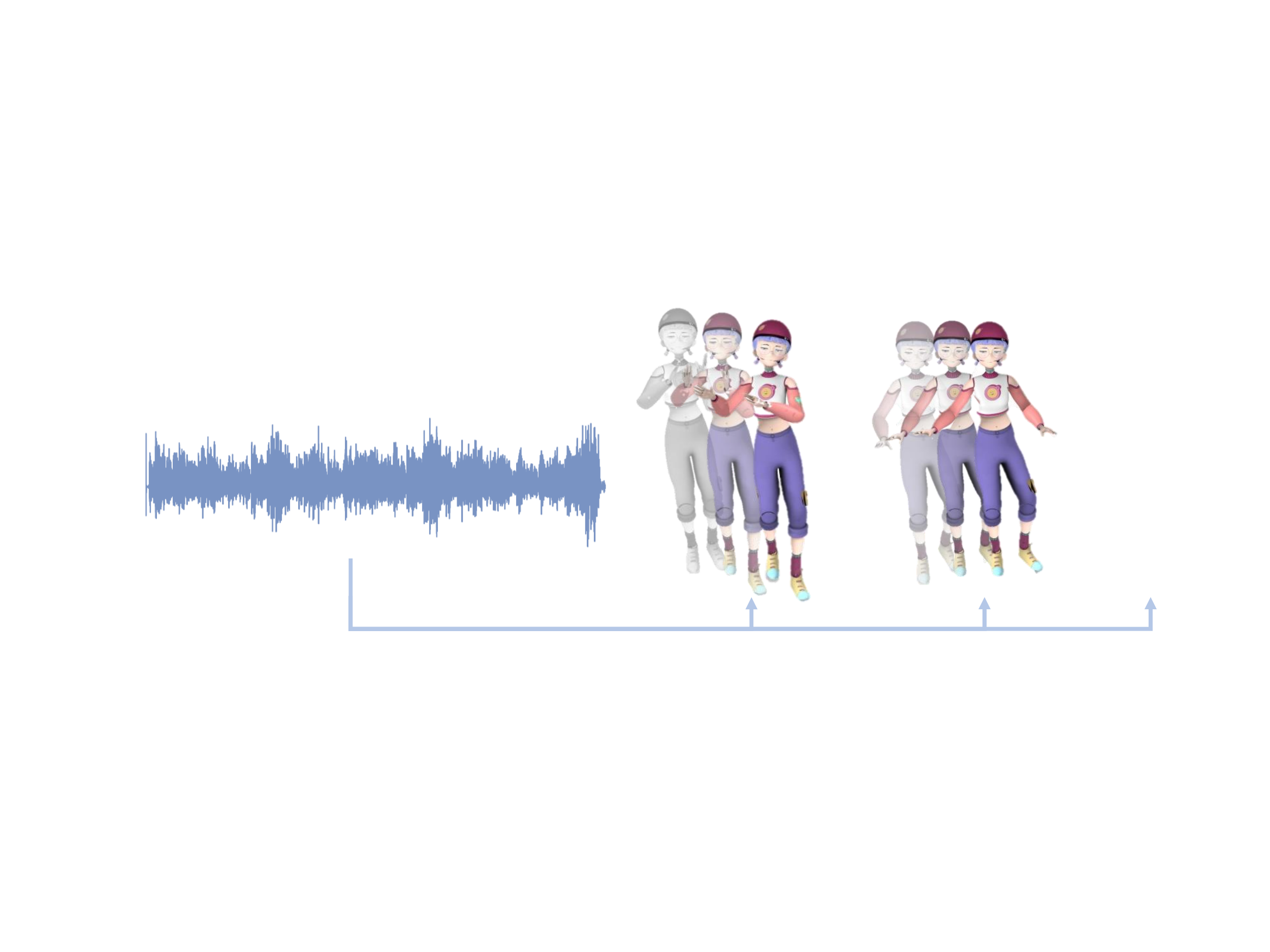}
\put(10,25){\small ``Completely"}
\put(50,35){\small Motion 1}
\put(72,35){\small Motion 2}
\put(91.5,35){\small Other}
\put(90,31){\small motions}
\put(95,15){$\cdots$}
\end{overpic}
\vspace{-6mm}
\caption{Illustration of the existence of one-to-many mapping between audio and motion in Trinity dataset~\cite{IVA:2018}. Different gestures are performed when the subject says ``completely''.Similar phenomena broadly exist in co-speech gestures. The character used for demonstration is from Mixamo~\cite{mixamo}.}
\label{demo}
\end{figure}

Existing algorithms developed for audio to body dynamics have some obvious limitations.
For example, \cite{speech2gesture} adapts a fully convolutional neural network to co-speech gesture synthesis tasks.
Nevertheless, their model tends to predict averaged motion and thus generates motions lacking diversity.
This is due to the underlying one-to-one mapping assumption of their model, which ignores that the relationship between speech and co-speech gesture is one-to-many in nature.
Under such an overly simplified assumption, the model has no choice but to learn the averaged motion when several motions match almost the same audio clips in order to minimize the error. 
The above evidence inspires us to study whether or not explicitly modeling this multimodality improves the overall motion quality.
To enhance the regression capability, we introduce an extra motion-specific latent code. 
With this varying {\it full} latent code, which contains the same shared code and varying motion-specific code, the decoder can regress different motion targets well for the same audio, achieving one-to-many mapping results.
Under this formulation, the shared code extracted from audio input serves as part of the control signal. 
The motion-specific code further modulates the audio-controlled motion, enabling multimodal motion generation.

\par
Although this formulation is straightforward, it is not trivial to make it work as expected.
Firstly, there exists an easy degenerated solution since the motion decoder could utilize only the motion-specific code to reconstruct the motion.
Secondly, we need to generate the motion-specific code since we do not have access to the target motion during inference.
Our solution to the aforementioned problems is providing {\it random noise} to the motion-specific code so that the decoder has to utilize the deterministic information contained in the shared code to reconstruct the target.

\par
But under this circumstance, it is unsuitable for forcing the motion decoder to reconstruct the exact original target motion anymore.
So a \textit{relaxed motion loss} is proposed to apply to the motions generated with random motion-specific code.
Specifically, it only penalizes the joints deviating from their targets larger than a threshold.
This loss encourages the motion-specific code to tune the final motion while respecting the shared code's control.

% \par
% A preliminary version of this work was presented in \cite{audio2gestures}.
% In this paper, we extend it from the following aspects:
% (1) We further analyze the predicted motion with more structured/higher-level metrics (\textit{i.e.} STFT, SSIM, LPIPS, and FID).
% % by introducing more metrics, including STFT, SSIM, LPIPS, and FID.
% % Since the tested structured loss  complementary to 
% We find them complementary to the previous point-wise metrics (\textit{e.g.} PCK) since they additionally consider local temporal and/or spatial structures.
% Then we test using them as training losses and obtain motions with improved quality, especially for STFT.
% % Noticing that there are some promising merits in the newly introduced metrics, we further take the metrics as a training loss to improve the Audio2Gestures algorithm.
% (2) We show our formulation is compatible with DCT space modeling, which enables the user to control the dynamics of generated motion by only preserving certain DCT components. 
% And we find the audio-motion shared code is more related to motion speed and rhythm while motion-specific code is more related to small range variations. \dknote{need check this conclusion}
% (3) We show our formulation is applicable to other backbones (\textit{i.e.} GRU and Transformer) besides TCN.
% (4) We conduct an extensive ablation study on the hyper-parameter of the relaxed motion loss using Trinity Gesture dataset.
% % to analyze the influence of the hyper-parameter of the relaxed motion loss. 
% % \dknote{update later. user study for the new motion losses are more important.}

\par
A preliminary version of this work was presented in \cite{audio2gestures}.
In this paper, we extend it from the following aspects:
% \bao{
(1) We thoroughly investigate structured and perceptual metrics (\textit{i.e.} STFT, SSIM, LPIPS, and FID) as training losses in our framework.
% } 
%further analyze the predicted motion with more structured/higher-level metrics (\textit{i.e.} STFT, SSIM, LPIPS, and FID).
% by introducing more metrics, including STFT, SSIM, LPIPS, and FID.
% Since the tested structured loss  complementary to 
We find them complementary to the previous point-wise losses since they additionally consider local temporal and/or spatial data structures. 
% \bao{
Especially with STFT, our network can consistently yield higher-quality motions.
% }
%Then we test using them as training losses and obtain motions with improved quality, especially for STFT.
% Noticing that there are some promising merits in the newly introduced metrics, we further take the metrics as a training loss to improve the Audio2Gestures algorithm.
% \bao{
(2) We improve the user controllability by switching the latent space into discrete cosine transform (DCT) space, where different DCT components can control different dynamics of the generated motions.
% }
%show our formulation is compatible with DCT space modeling, which enables the user to control the dynamics of generated motion by only preserving certain DCT components. 
And we find the audio-motion shared code is more related to motion speed and rhythm while motion-specific code is more related to small range variations. 
% \dknote{need check this conclusion}
% \bao{
(3) We conduct extensive ablation studies on different network backbones and hyper-parameters of motion losses and present more detailed discussions and analysis on the observations.
% }
%(3) We show our formulation is applicable to other backbones (\textit{i.e.} GRU and Transformer) besides TCN.
%(4) We conduct an extensive ablation study on the hyper-parameter of the relaxed motion loss using Trinity Gesture dataset.
% to analyze the influence of the hyper-parameter of the relaxed motion loss. 
% \dknote{update later. user study for the new motion losses are more important.}

\par
The overall contributions are summarized as follows:
\begin{compactitem}
\item We present a co-speech gesture generation model whose latent space is split into audio-motion shared code and motion-specific code, to better model the training data pairs and generate diverse motions.
% 
% \item We utilize random sampling and a relaxed motion loss to better avoid degeneration of the proposed network and enable multimodal motion generation given the same     audio input.
\item We propose a new relaxed motion loss, accompanied by other training losses/strategies, to better avoid degeneration of the proposed network, enabling multimodal motion generation given the same audio input.
\item The effectiveness of the proposed method has been verified on 3D and 2D gesture generation tasks by comparing it with several state-of-the-art methods. And this split formulation is compatible with DCT space modeling and other backbones (\textit{i.e.} GRU, Transformer), and robust to hyper-parameter choice in the relaxed motion loss.
\item In complement to the most commonly used point-wise metrics/losses, we analyze the generated motion by taking local spatial and/or temporal structure into consideration and introduce them as training losses to further improve the motion quality.
% effectiveness of the proposed method from multiple aspects with different metrics.
% \item \dk{We show our formulation is compatible with DCT space modeling and other backbones (\textit{i.e.} GRU, Transformer), and robust to hyper-parameter choice in the relaxed motion loss.}
% \item \lj{We show the audio-motion shared code and the motion-specific code could be modeled either in temporal domain and frequency domain. Further more, modeling the codes in frequency domain enables control the frequency of the generated motion.}
% 
% \item \lj{Experiment shows that the framework is robust to the hyper-parameter of the relaxed motion loss and different backbones.}

% \item The framework is analysed in detail from multiple aspects with different metrics.
\item As a byproduct, the proposed method is suitable for motion synthesis from annotations since it can well respect the predefined actions on the timeline by simply using their corresponding motion-specific code.
% \item Experiments show that the framework is robust to the backbone of the encoder and the hyper parameter.
\end{compactitem}

\section{Related work}

\noindent\textbf{Audio to body dynamics.}
Early motion generation methods usually blend \emph{motion clips/segments} chosen from a motion database according to hidden Markov model~\cite{Levine2009} or conditional random fields~\cite{gesture_controllers}. 
Algorithms focusing on selecting motion candidates from a pre-processed database usually cannot generate motions out of the database and does not scale to large databases.
% \lj{
Recently, \cite{motion_matching} extends this motion-matching based framework with a deep post-processing, where the retrieved motion clip using kNN from the database is used as input to a conditional GAN to generate more diverse results.
% provides users controllability by proposing a motion-matching based method, which first selecting a possible motion clip with kNN from a motion database, and then improving the synthesised motion quality with a conditional GAN.
% }
% \lj{
% Another segment-based method, 
Rhythmic Gesticulator~\cite{ao2022rhythmic}, which is also segment-based, generates gesture clip-by-clip via a VQ-VAE network to achieve better motion rhythm since its generator network is trained with motion clips segmented according to motion beats.
% improves the rhythm of the generated gestures by first segmenting the motions to clips according the audio beats and modeling the motions within each clips.
% }

\par
Recently, deep generative models (e.g. VAEs~\cite{vae} and GANs~\cite{gan}) have achieved great success in generating realistic images, as well as human motions~\cite{mtvae,motioninpainting,motionvae}.
For example,~\cite{audio2body_dynamics} utilizes a classic LSTM to predict the body movements of a person playing the piano or violin given the the instrumental music. 
However, the body movements of a person playing the piano/violin show regular cyclic pattern and are usually constrained within a small pose space.

\par
In contrast, generating co-speech gestures is more challenging in the following two aspects -- the motion to generate is more difficult and the relationship between the speech and motion is more complicated.
As a result, more powerfully networks are proposed and they often are trained with more data.
% to tackle this task.
For example,
% As a result, 
Speech2Gesture~\cite{speech2gesture} proposes a more powerful fully convolutional network, consisting of a 8-layer CNN audio encoder and a 16-layer 1D U-Net decoder, to translate log-mel audio feature to gestures.
HA2G~\cite{ha2g} proposes to generate fine-grained gestures by associating the hierarchical feature between audio and motion.
Speech2Gesture is 
% And this network is 
trained with 14.4 hours of data per individual on average in comparison to 3 hours data in~\cite{audio2body_dynamics}.
Other than greatly enlarged network capacity, the fully convolutional network also better avoids the error accumulation problem often faced by RNN-based methods.
However, it still suffers from predicting the averaged motion due to the existence of one-to-many mapping in the training data.
The authors further introduce adversarial loss and notice that the loss helps to improve diversity but degenerates the realism of the outputs.

In contrast, our method avoids learning the averaged motion by explicitly modeling the one-to-many mapping between audio and motion with the help of the extra motion-specific code.
Similar to our split latent space formulation, DisCo~\cite{disco} uses contrastive learning to disentangle the motion feature into motion rhythm and motion content.

\par
Due to the lack of 3D human pose data, the above deep learning based methods~\cite{audio2body_dynamics, speech2gesture} have only tested 2D human pose data, which are 2D key point locations estimated from videos.
Recently,~\cite{IVA:2018} collects a 3D co-speech gesture dataset named Trinity Speech-Gesture Dataset, containing 244 minutes of motion capture (MoCap) data with paired audio, and thus enables deep network-based study on modeling the correlation between audio and 3D motion. 
This dataset has been tested by StyleGestures~\cite{moglow}, which is a flow-based algorithm~\cite{glow, moglow}. StyleGestures generates 3D gestures by sampling poses from a pose distribution predicted from previous motions and control signals.
However, samples generated by flow-based methods~\cite{glow, moglow} are often not as good as VAEs and GANs.
In contrast, our method learns the mapping between audio and motion with a customized VAE.
Diverse results can be sampled since VAE is a probabilistic generation model.
% \dknote{check}

\input{img/fig_network}

\noindent\textbf{Human motion prediction.}
There exist many works focus on predicting future motion given previous motion~\cite{motioninpainting,quaternet,mtvae}.
It is natural to model sequence data with RNNs~\cite{ERD, Structural-RNN,martinez2017human, mtvae}.
But~\cite{motioninpainting} has pointed out the RNN-based methods often suffer from error accumulation and thus are not good at predicting long-term human motion. 
So they proposes to use a fully convolutional generative adversarial network and achieves better performance at long-term human motion prediction.
Similarly, we also adopt a fully convolutional neural network since we need to generate long-term human motion.
Specific to 3D human motion prediction, another type of error accumulation happens along the kinematic chain~\cite{quaternet} because any {\it small} joint rotation error propagates to all its descendant joints, \textit{e.g.} hands and fingers, resulting in {\it considerable} position error especially for the end-effectors (wrists, fingers).
So QuaterNet~\cite{quaternet} optimizes the joint position which is calculated from forward kinematics when predicting long-term motion.
Differently, we optimize the joint rotation and position losses at the same time to help the model learn the joint limitation at the same time.

\noindent\textbf{Multimodal generation tasks.}
Generating data with multimodality has received increasing interests in various tasks, such as image generation~\cite{MUNIT, bicyclegan}, motion generation~\cite{MoCoGAN, S3VAE}.
For image generation, MUNIT~\cite{MUNIT} disentangles the embedding of images into content feature and style feature. 
BicycleGAN~\cite{bicyclegan} combined cVAE-GAN~\cite{vaegan} and cLR-GAN~\cite{InfoGAN,bigan} to encourage the bijective consistency between the latent code and the output so that the model could generate different output by sampling different codes. 
For video generation, MoCoGAN~\cite{MoCoGAN} and S3VAE~\cite{S3VAE} disentangle the motion from the object to generate videos in which different objects perform similar motions.
Different from~\cite{MoCoGAN,S3VAE}, our method disentangle the motion representation into the audio-motion shared information and motion-specific information to model the one-to-many mapping between audio and motion.

\section{Network overview} \label{sec:overview}

The whole pipeline of our method is illustrated in \reffig{fig:model}.
It is a conditional variational autoencoder (cVAE), with audio features as the condition (\textit{i.e.} control signal) to generate co-speech gestures.
Given an audio-motion pair $\{A, M\}$, the latent code used to reconstruct the motion clip has been explicitly {\it split} into two parts (\textit{i.e.} \textit{shared} code $S$ and \textit{motion-specific} code $I$) to account for the frequently occurred \textit{one-to-many mapping} between the {\it same} (technically, very similar) audio and many different possible motions.

Under this formulation, given the same audio input (resulting in the same shared code $S_\text{A} = f_\text{A}(A)$), varied motions produce different motion-specific code $I_\text{M}$ through motion encoder $f_\text{M}$ (i.e. $(S_\text{M}, I_\text{M}) = f_\text{M}(M)$), resulting in different {\it full} latent codes ($S_\text{A}\oplus I_\text{M}$) so that the network can better model their relation from the audio-motion training pairs ($\hat{M} = g(S_\text{A}, I_\text{M})$).
% one-to-many mapping ($\hat{M} = g(S_\text{A}, I_\text{M})$).

Since we do not have the target motion to extract $I_M$ at inference time, 
a variational autoencoder~\cite{vae} that allows latent space sampling is adopted to obtain a suitable motion-specific code $I$.
During inference, shared feature $S_\text{A}$ 
% \dk{(calibrated with motion feature \dknote{insert, also ref to this loss})} 
is extracted with $f_\text{A}$ from the given audio $A$, serving as the control signal.
Motion-specific feature $I_\text{R}$ is generated with $f_\text{R}$ from a randomly sampled signal.
An additional mapping network is introduced empirically to facilitate its generation process.
% Both $S_\text{A}$ and $I_\text{R}$ are fed into the decoder $g$ to produce the final motion $M$, i.e $M = g(S_\text{A}, I_\text{R})$.
Finally, The co-speech gesture is generated by the decoder $g$ that takes both $S_\text{A}$ and $I_\text{R}$ as input.

%%%%%%%%%%%%%%%%%%%%%%%
%%%%%%%% Network %%%%%%
%%%%%%%%%%%%%%%%%%%%%%%
% \section{Network}
% \lj{In this section, we discuss the network structure of the proposed Audio2Gestures algorithm.}
% \lj{The proposed Audio2Gestures algorithm have little limitation on the network structure, except that the motion encoder $f_m$ have two heads to split the motion feature into audio-motion shared feature $S_M$ and motion specific feature $I_M$.
% In this section, we describe two instances of the Audio2Gestures, including a Variational AutoEncoder (VAE) based model, named Audio2Gestures-VAE, and a Vector Quantised-Variational AutoEncoder (VQ-VAE) based model, named Audio2Gestures-VQVAE.}
% \subsection{Variational autoencoder}
% Compared to autoencoder, VAE additionally imposes constraints on the latent code to enable sampling outputs from the latent space.

% \subsection{Vector Quantised-Variational AutoEncoder}
% \lj{VQ-VAE have received many attention due to the model alleviate the mode-collapse problem of VAE. The model generate samples by sampling from a learned dictionaries with a prior.}
\section{Network training}
% In this section, we detail the training process, including our training strategies and training losses of the proposed Audio2Gestures algorithm.
In this section, we detail the training process, including commonly used newly introduced motion losses in Sec.~\ref{sec:motion_losses} to help the network generate a valid pose, 
and our latent space learning losses/strategies in Sec.~\ref{sec:latent_learning} to help the network to decouple the latent code into audio-motion shared and motion-specific code.

\subsection{Motion learning}
\label{sec:motion_losses}

In this section, we will describe the losses introduced for better motion reconstruction.
% The motion reconstruction loss (\refsec{sec:motion_recon_loss}) is the most basic loss, which optimizing the joint rotation, joint position and joint speed at the same time.
% Basic losses for motion reconstruction typically include optimizing joint rotation, joint position, and joint speed at the same time.
Our basic motion reconstruction loss (Sec.~\ref{sec:motion_recon_loss}), which includes 3 most typically used individual loss terms, \textit{i.e.} joint rotation, joint position, and joint speed losses at the same time.
However, the motion reconstruction loss is a \textit{point-wise} loss, which means the predicted motion is optimized frame-to-frame and joint-by-joint.
This point-wise loss pays little attention to the motion dynamics,  thus easily resulting in averaged motion \cite{speech2gesture}.
To help the network better model the motion dynamics, 
losses considering per-joint temporal dynamics (\textit{i.e.} STFT in~\refsec{sec:stft} and SSIM in~\refsec{sec:ssim}) and a loss measuring abstract high-level similarity across different joints (\textit{i.e.} LPIPS in~\refsec{sec:lpips}) are introduced as a complement to the typical motion reconstruction loss.
Their influences on the final generated motions are discussed in \refsec{sec:result:new_motion_losses}.
% STFT loss and SSIM loss is introduced, which is described in \refsec{sec:stft} and \refsec{sec:ssim}, respectively.
% 
% But above losses still optimizing the predicted motion joint-by-joint and do not take the correlation between the joints into account. Thus, LPIPS~\cite{lpips} is introduced, which supervises the generated motion within the deep motion space. 

\subsubsection{Motion reconstruction loss}
% \noindent\textbf{Motion reconstruction loss}
\label{sec:motion_recon_loss}
We adopt the most commonly used motion losses as our basic \textit{motion reconstruction loss}, including rotation loss, position loss, and speed loss.
% In our method, the generated motion is supervised with {\it motion reconstruction loss}, consisting of rotation loss, position loss, and speed loss. 
Formally, it is defined as follows:
\begin{equation}
% \small
% \Scale[1]{
    L_{\text{mot}} = \lambda_{\text{rot}} \times L_{\text{rot}} + \lambda_{\text{pos}} \times L_{\text{pos}} + \lambda_{\text{speed}} \times L_{\text{speed}}
% }, 
\label{eq:mr}
\end{equation}
where $\lambda_{\text{rot}}$, $\lambda_{\text{pos}}$, $\lambda_{\text{speed}}$ are weights. We detail each term in the following.

Angular distance, \textit{i.e.}, geodesic distance, between the predicted rotation and the GT is adopted as the rotation loss.
Mathematically,
\begin{equation}
\Scale[1]{
    L_{\text{rot}} = \frac{1}{J\times T}\sum_{j=1}^J\sum_{t=1}^T cos^{-1}\frac{\operatorname{Tr}{\big(R_t^j(\hat{R}_t^j)^{-1}\big)} - 1}{2}
}
\end{equation}

Position loss is the $L_1$ distance between the predicted and target joint positions as follows:
\begin{equation}
\Scale[1]{    
    L_{\text{pos}}= \frac{1}{J\times T}\sum_{j=1}^J\sum_{t=1}^T\Vert\,\hat{p}_t^j-p_t^j\,\Vert_1
}
\end{equation}

Speed loss is introduced to help the model learn the complicated motion dynamics. 
In our work, the joint speed $v^j_t$ is defined as $v^j_t=p^j_{t+1} - p^j_t$.
We optimize the predicted and target joint speed as follows:
\begin{equation}
\Scale[1]{        
    L_{\text{speed}} = \frac{1}{J\times (T-1)}\sum_{j=1}^J\sum_{t=1}^{T-1}\Vert\,\hat{v}_t^j - v_t^j\,\Vert_1
}
\end{equation}

Our model can be trained with 2D motion data or 3D motion data.
When modeling the 2D human motion, our method directly predicts the joint position.
When modeling the 3D human motion, our method predicts the joint rotation and calculates the 3D joint positions with forward kinematics (FK).
Concretely, the FK equation takes in as input the joint rotation matrix about its parent joint and the relative translation to its parent joint (\textit{i.e.} bone length) and outputs joint positions as follows:
\begin{equation}
% \small
    p^j_t = p^{\text{parent}(j)}_t + R^j_ts^j
\label{eq:fk}
\end{equation}
where $R^j_t$ represents the rotation matrix of joint $j$ in frame $t$,
$p^j_t$ represents the position of joint $j$ in frame $t$,
$s^j$ represents the relative translation of joint $j$ to its parent,
and $\text{parent}(j)$ represents the parent joint index of the joint $j$.
We will always use $j$ and $t$ to index joints and frames in the following.
Our model predicts joint rotation in 6D representation~\cite{6d_repr}, which is a continuous representation that help the optimization of the model.
The representation is then converted to rotation matrix $R^j_t$ by Gram-Schmidt-like process, where $R^j_t$ is the rotation matrix of joint $j$ in frame $t$.

\subsubsection{STFT} 
% \noindent\textbf{\lj{STFT}}
\label{sec:stft}
% To compare the motion dynamic between generated motion and the groundtruth, we introduce STFT metric.
To help the network learn better temporal motion dynamics, we introduce Short-term Fourier transform (STFT) loss as a supervision signal. 
Following \cite{denoiser}, the metric is calculated by:
% \dknote{add STFT parameters here}
\begin{equation}
\label{eq:stft}
    L_{\text{stft}}(p,\hat{p})=\frac{1}{T}\Vert \mathop{\log}|STFT(p;w,s)|-\mathop{\log}|(STFT(\hat{p};w,s)|\Vert_1,
\end{equation}
where $w$ is the Hanning window of size 32 and the stride size $s$ is set to $1/4$ of the window size.
% \dk{A hanning window of size \dknote{insert} and the stride size is $1/4$ of the window size.}

\subsubsection{SSIM} 
% \noindent\textbf{\lj{SSIM}}
\label{sec:ssim}
Structural similarity index measure (SSIM)~\cite{ssim}, which is originally proposed to measure the similarity between two images, considers local inter-dependencies among pixels.
% which is originally proposed to measure the image quality by comparing luminance, contrast, and structure similarity between two images.
By treating joint number and frame dimensions as ``image'' height and width and xyz as channels, we can define the SSIM between two motions to measure their similarity with the consideration of local inter-dependencies (\textit{i.e.} motion dynamics).
Concretely, motion SSIM and the corresponding luminance $l$ (\textit{i.e.} averaged joint position), contrast $c$, and structure $s$ are defined as follows.
% 
% We adopt the metric to measure the quality the motion by compare the similarity of the averaged joint position $l$, joint dynamics $c$ and motion structure similarity $s$.
\begin{equation}
\label{eq:ssim}
    SSIM(p,\hat{p}) = [l(p,\hat{p})]^\alpha\cdot[c(p,\hat{p})]^\beta\cdot[s(p,\hat{p})]^\gamma,
\end{equation}
% where $l$, $c$ and $s$ measures the luminance, contrast and structure similarity of image $x$ and $y$, which are calculated by:
where $l$, $c$ and $s$ are calculated by:
\begin{gather}
l(p,\hat{p})=\frac{2\mu_p\mu_{\hat{p}}+C_1}{\mu_p^2+\mu_{\hat{p}}^2+C_1} \\
c(p,\hat{p})=\frac{2\sigma_p\sigma_{\hat{p}}+C_2}{\sigma_p^2+\sigma_{\hat{p}}^2+C_2} \\
s(p,\hat{p})=\frac{\sigma_{p\hat{p}}+C_3}{\sigma_p\sigma_{\hat{p}}+C_3}
\label{eq:ssim_lcs}
\end{gather}
% \begin{equation}
%     c(p,\hat{p})=\frac{2\sigma_p\sigma_{\hat{p}}+C_2}{\sigma_p^2+\sigma_{\hat{p}}^2+C_2}
% \end{equation}
% \begin{equation}
%     s(p,\hat{p})=\frac{\sigma_{p\hat{p}}+C_3}{\sigma_p\sigma_{\hat{p}}+C_3}
% \end{equation}
% 
The $\mu_p$ ($\mu_{\hat{p}}$) is the average of joint positions across time:
% of a joint, which is calculated by averaging all the values across time:
\begin{equation}
    \mu_p = \frac{1}{N}\sum_{t=1}^{T}p_t.
\end{equation}
The $\sigma_{p}$
% and $\sigma_{\hat{p}}$ is the dynamic, which 
is the standard deviation of the joint positions across time:
\begin{equation}
    \sigma_p = (\frac{1}{N-1}\sum_{t=1}^T(p_t-\mu_p)^2)^{\frac{1}{2}}.
\end{equation}
The $\sigma_{xy}$ is the covariance of the joint positions across time:
\begin{equation}
    \sigma_{p\hat{p}} = \frac{1}{N-1}\sum_{t=1}^{T}(p_i-\mu_p)(\hat{p}_i-\mu_{\hat{p}})
\end{equation}
In our experiments, the $\alpha$, $\beta$ and $\gamma$ are all set to 1.
$C_1$, $C_2$ and $C_3$ are set as $0.01^2$, $0.03^2$ and $\frac{0.03^2}{2}$, respectively.

\subsubsection{LPIPS} \label{sec:lpips}
% LPIPS~\cite{lpips} measures the similarity of the generated motion with the groundtruth in the deep feature space.
Learned perceptual image patch similarity (LPIPS)~\cite{lpips} measures the similarity between the generated motion and the groundtruth in feature space, which focuses on abstract high-level motion characteristics in complement to low-level joint-wise error, and has been used in many other generation tasks~\cite{stargan_v2}.
\begin{equation}
\label{eq:lpips}
    L_{\text{perceptual}} = \sum_l\frac{1}{T}\sum_{t=1}^T\Vert f^l_t-\hat{f}_t^l \Vert_2^2
\end{equation}
where $f_t^l$ is the deep feature produced by the $l^{th}$ residual block at $t^{th}$ frame of the input motion.
 
Concretely, the network used to extract deep motion features is a convolutional autoencoder
% To extract the deep motion features, a motion autoencoder is 
trained on Trinity Gesture dataset~\cite{IVA:2018} using motion reconstruction loss in Eq.~\ref{eq:mr}.
The encoder and the decoder of the model consist of 5 residual blocks each~\cite{resnet} and take 6D joint rotation~\cite{6d_repr} as input and output.
One residual block consists of two 1D dilated convolution layers. 
The final receptive field size of the motion encoder is 125 (frames).

%%%%%%%%%%%%%%%%%%%%
%% Audio2Gestures %%
%%%%%%%%%%%%%%%%%%%%

\input{img/fig_training}

\subsection{Latent code learning}  
\label{sec:latent_learning}

The shared code is expected to model the strong correlation between audio and motion while the motion-specific code is expected to capture diverse motion information more independent of the audio.
So we need to 1) prevent any of $S$ and $I$ is ignored by the decoder (i.e. degenerate) and 2) encourage shared code $S_A$ to contain more information and can be used interchangeably with $S_M$ so that the audio input can effectively affect the output motion.

Shared code $S$ is easily ignored by the motion decoder since motion AE is an easier task than cross modality translation.
% But in practice, we observe the motion decoder easily ignores the shared codes $S$ and reconstructs the motion only from $I_\text{M}$.
% This is unwanted since the final motion is solely determined by the motion-specific features, being completely not correlated with the control signal (audio).
%  
Thus, another data flow ($\hat{M}_{S_\text{A}I_\text{R}}=g(S_\text{A}, I_\text{R})$, where $I_\text{R}$ is from random noise.) is introduced so that the decoder has to utilize the information contained in the shared code extracted from audio to reconstruct the target.
% 
% The $I_\text{R}$ is generated from the mapping net $f_\text{R}$, whose input is a random signal from a Gaussian distribution.
% The mean and variance of the distribution are calculated from the $I_\text{M}$ of the target motion per channel during training.

% \subsection{Training losses}
% According to different purposes, the training losses can be categorized into two groups, including motion learning losses, which help the network to generate a valid pose, 
% and latent code learning losses, which help the network to decouple the latent code into audio-motion shared and motion-specific code.
% \dknote{merged into this section. check later}

% To learn better split audio-motion shared code and motion-specific latent code, 
In total, five types of losses are introduced in addition to previous motion loss, 
including a shared-code \textit{alignment constraint}, a \textit{KL divergence (omitted in the figure)}, a \textit{relaxed motion loss}, a \textit{bicycle constraint}, and a \textit{diversity loss}
(see \reffig{fig:training}).
% 
% Aside from the basic motion reconstruction loss in Eq.~\ref{eq:mr},
The {\it alignment constraint} and {\it relaxed motion loss} contribute to better learned joint embedding (\textit{i.e.}, shared code) of the audio and motion.
% are introduced to better learn the joint embedding (\textit{i.e.}, shared code) of the audio and motion. 
The {\it relaxed motion loss} and {\it bicycle constraints} help avoiding degeneration (\textit{i.e.} some part of the latent code is ignored and loses controllability on the final generated motion).
The {\it bicycle constraints} and {\it diversity loss} are introduced to better model the multimodality of the motions.
% 
% The \lj{{\it KL divergence} constrains the latent space of the motion-specific code to help sampling the motion-specific feature at inference time.}
% 
Detailed descriptions are in the following.
% \dknote{updated. check}
% \dknote{need update, 6 losses in total}

\subsubsection{Shared code alignment} \label{sec:latent:align}
The shared code of paired audio and motion is expected to be the same so that we can safely use audio-extracted shared code during inference and generate realistic and audio-related motions.
We align the shared code of audio and motion by the following alignment constraint:
\begin{equation}
    L_\text{AC}= \Vert\, S_\text{A} - S_\text{M} \,\Vert_1. \label{eq:sca}
\end{equation}

\subsubsection{KL divergence}
During training, the distribution $P$ of the latent code is constrained to match a target distribution $Q$ with KL divergence as follows: 
\begin{equation}\small
\Scale[0.9]{\mathcal{D}\big(Q(z) \,\Vert\, P(z|X)\big) =  E_{z\sim Q}\big[\log{Q(z)}-\log{P(z|X)}\big]} ,
\end{equation}
where the $X$ represents the input of the corresponding encoder (audio or motion in our case), and $z$ represents its corresponding latent code.
The above goal can be achieved by minimizing the Evidence Lower Bound (ELBO) \cite{vaetutorial}:
\begin{equation}\small
\Scale[0.9]{\log{P(X|z)}-\mathcal{D}\big[Q(z|X)\,\Vert\, P(z)\big].} \label{eq:target}
\end{equation}

The second term of Eq.~\ref{eq:target} is a KL-divergence between two Gaussian distributions (with a diagonal covariance matrix). The prior distribution $P$ is set to Gaussian distribution (with a diagonal covariance matrix in our model, thus, the KL-divergence can be computed as:
\begin{equation}\small
\Scale[0.9]{\mathcal{D}=\frac{1}{2}\Big(\operatorname{tr}\big(\Sigma(X)\big)+\mu(X)^T\mu(X)-k-\log{\det\big(\Sigma(X)\big)}\Big)},
\end{equation}
where $k$ is the dimension of the distribution~\cite{vaetutorial}.

\subsubsection{Degeneration avoidance}
% As we described in \refsec{sec:overview}, 
The motion decoder easily ignores the shared code input and solely use the motion-specific code to reconstruct the output motion (\textit{i.e.} a motion auto-encoder)
since it is a relatively easier task compared to audio2gestures that requires \textit{translating} from one modality to another.
This solution is a \textit{degenerate} network, where the audio encoder is wasted and cannot control the final motion.
% the motion decoder easily results in the degenerated network, which means the shared code is completely ignored and has no effect on the generated motion. 
Our solution to alleviating such degeneration is introducing an extra motion reconstruction task which takes shared code $S_\text{A}$ extracted from the audio but random motion-specific code $I_R$ as input.
Ideally, the generated motion $\hat{M}_{S_\text{A}I_\text{R}}$ resembles its GT in some aspects but is not identical to its GT. 
In our case, we assume the generated poses are similar in the 3D world space.
Thus we propose {\it relaxed motion loss}, which calculates the position loss and penalizes the model only when the distance is larger than a certain threshold $\rho$:
\begin{equation}
% \Scale[0.9]{    
L_\text{S} = \frac{1}{J} \sum_{i=1}^J \max\big(\Vert\,\hat{p}_i - p_i \,\Vert_1-\rho, 0\big). 
\label{eq:rmr}
% }
\end{equation}

% \subsubsection{Motion-specific code reconstruction} 

Recall that, to better model the relation between audio-motion pairs, we expect the existence of one-to-one correspondence (\textit{i.e.} bijection) between the \textit{full} latent code and the motion by introducing split latent spaces (\textit{i.e.} shared latent code and motion-specific latent code).
To encourage a bijection during training, we introduce bicycle constraint~\cite{bicyclegan} (\textit{i.e.} $M \rightarrow I \rightarrow M$ and $I \rightarrow M \rightarrow I$) to handle this mode-collapse problem.
% 
% \dk{By explicitly splitting the full latent code into two parts to handle the one-to-many mapping issue, the network could generate multimodal motions (\textit{i.e.} motions in different types) using different but appropriate motion-specific codes.}
% % 
% % Although the network could model the multimodal distribution of audio-motion pair by splitting the motion code into audio-motion shared one and motion-specific ones, it is not guaranteed the decoder can sample multimodal motions. 
% % 
% For example, suppose the mapping net only maps the sampled signal to a single mode of the multimodal distribution. 
% % 
% In that case, the decoder still could only generate unimodal motions, which is also known as the mode-collapse problem.
% 
% The bicycle constraint~\cite{bicyclegan} ($M \rightarrow I \rightarrow M$ and $I \rightarrow M \rightarrow I$) is introduced to avoid the mode-collapse problem, which encourages a bijection between the motion and the motion-specific code.
Since the motion reconstruction loss has already been introduced, an extra reconstruction loss of the motion-specific code is introduced as supplement:
\begin{equation}\label{eq:cr}
% \Scale[0.9]{    
L_\text{cyc} = \Vert\, \hat{I}_\text{R} - I_\text{R} \,\Vert_1 
% }.
\end{equation}

\subsubsection{Motion-specific code generation}

By explicitly splitting the full latent code into two parts to handle the one-to-many mapping issue, the network could generate multimodal motions (\textit{i.e.} motions in different types) using different and \textit{appropriate} motion-specific codes.
Although the motion-specific code could be sampled from Gaussian distribution directly, we noticed that the realism and diversity of the generated motions are not good.
We hypothesize this is caused by the misalignment of the Gaussian distribution and the motion-specific code distribution.
Thus, a mapping network is introduced to transform the signal sampled from Gaussian distribution to the motion-specific embedding.
Concretely, we calculate the mean and variance for every channel and every sample of the $I_\text{M}$ at the training stage.
The sampled features will be fed into a mapping network, which is also a variational autoencoder, before concatenating them with the shared code to generate the final motions.

\subsubsection{Motion diversification}
To further encourage multimodality of the generated motion, a diversity loss~\cite{MSGAN,stargan_v2} is introduced.
\begin{equation}
\label{eq:ds}
% \Scale[0.9]{
L_\text{DS}=-L_{\text{pos}}(\hat{M}_{S_{\text{M}}I_{\text{R}_1}}, M).
% }
\end{equation}
Maximizing the diversity of the generated motions encourages the mapping network to explore the meaningful motion-specific code space. 
We follow the setting in~\cite{stargan_v2} and directly maximize the joint position distance between two sampled motions since it is more stable than the original one~\cite{MSGAN}.

%%%%%%%%%%%%%%%%%%%%%
%%%% Experiments %%%%
%%%%%%%%%%%%%%%%%%%%%
\section{Datasets, evaluation, \& implementation details}

In this section, we first introduce the datasets, evaluation metrics and implementation details separately in \refsec{sec:datasets}-\ref{sec:implementation}. 
Then we show the performance of our algorithm and compare it with three state-of-the-art methods \ref{sec:sota}. 
Finally, we analyze the influence of each module of our model on the performance by ablation studies \ref{sec:ablation}.
More results are presented in our project page\footnote{\url{https://jingli513.github.io/audio2gestures}}.

\subsection{Datasets} \label{sec:datasets}

\noindent\textbf{Trinity dataset.}
Trinity Gesture Dataset~\cite{IVA:2018} is a large-scale speech to gesture synthesis dataset. 
This dataset records a male native English speaker talking many different topics, such as movies and daily activities. 
The dataset contains 23 sequences of paired audio-motion data, 244 minutes in total. 
% \dk{
We randomly split it into training set, which contains 19 sequences, and test set, which contains 4 sequences (i.e. NaturalTalking 01/25/27/30).
All the training sequence frames are used for training and only the first 5000 frames of a test sequence are used for evaluation.
% }
% \lj{The first 5000 frames of four sequences are randomly selected as test data (NaturalTalking 01/25/27/30) and the others are used as training data.} 
The audio of the dataset is recorded at 44kHz.
The motion data, consisting of 56 joints, are recorded at 60 frame per second (FPS) or 120 FPS using Vicon motion capture system.
% \dknote{add dataset split here}

\noindent\textbf{S2G-Ellen dataset.}
The S2G-Ellen dataset, which is a subset of the Speech2Gesture dataset~\cite{speech2gesture}, contains positions of 49 2D upper body joint estimated from 504 YouTube videos, including 406 training sequences (469513 frames), 46 validation sequences (46027 frames), and 52 test sequences (59922 frames).
The joints, which is estimated using OpenPose~\cite{openpose}, include neck, shoulders, elbows, wrists, and hands.

\subsection{Evaluation metrics}\label{sec:eval}

% \subsubsection{Quantitative metrics}
% \TODO{SSIM, LPIPS, STFT}
It is still an open problem to effectively measure the quality of the generated motions with objective metrics.
In this paper, we measure the audio-driven gesture generation algorithms from three aspects, including similarity (Sec.~\ref{sec:metric:similarity}), diversity (Sec.~\ref{sec:metric:diversity}), and multimodality ((Sec.~\ref{sec:metric:multimodality})).

% to \dknote{TODO. Do they (quantitative and qualitative metrics) align well? relate them in the result analysis section.}

\subsubsection{Similarity metrics} \label{sec:metric:similarity}

Although the correlation between speech audio and co-speech gesture is as strong as that in talking face, certain patterns have been observed especially for \textit{short-term} and \textit{personalized} speech gesture generation scenario. 
Thus, similarity metrics are often adopted as an indicator to measure if the expected gestures have been generated.

Following Speech2Gesture~\cite{speech2gesture}, the $L_1$ distance of joint position in Eq.~\ref{eq:l1} and the percentage of correct keypoints (PCK) in Eq.~\ref{eq:PCK} are adopted to evaluate the correlation between the generated motion and the input audio.

However, both $L_1$ and PCK are low-level point-wise losses computing similarity joint-by-joint and frame-by-frame, resulting in very different rankings from human perception.
% , which compare the similarity between the generated motion and the groundtruth frame-to-frame.
For example, it has been noticed that static motion (\textit{e.g.} mean pose of the dataset) sometimes could also achieve a good $L_1$ and PCK performance~\cite{speech2gesture}.
So we additionally introduce several structural metrics to consider spatial and temporal similarities as complements, including short-term Fourier transform (STFT), structural similarity index measure (SSIM)~\cite{ssim}, learned perceptual image patch similarity (LPIPS)~\cite{lpips}, and Fréchet inception distance (FID).

\vspace{2mm}
{\it $L_1$ distance} is the average joint position error between corresponding joints between prediction $\hat{p}$ and GT $p$:
\begin{equation}\label{eq:l1}
% \Scale[0.9]{    
L_1 = \frac{1}{T\times J}\sum_{t=1}^T \sum_{j=1}^J\Vert\hat{p}_t^j-p_t^j\Vert_1 
% }.
\end{equation}
% 
% However, the metric is not comparable if the joint position is calculated with different skeleton (Larger skeleton will result to larger $L_1$ error). 
Note that the metric is comparable across different methods only if the same skeleton is used (\textit{e.g.} larger skeleton corresponding to larger $L_1$ difference).
We recommend normalize according to wrist-to-wrist distance if different skeletons are used in different methods. For our skeleton, the wrist-to-wrist distance is 120.64 cm.
\begin{equation}\label{eq:l1_}
% \Scale[0.9]{    
L_1^* = L_1 *\frac{l^*}{l^{curr}} 
% }.
\end{equation}
where $l^{curr}$ is the wrist-to-wrist length of the used skeleton and $l^*$ is the wrist-to-wrist length of a reference skeleton.
% \sout{Thus, we also report a normalized $L_1$ metric, which normalize the $L_1$ score using the distance between two wrist joints in T-pose.}

\vspace{2mm}
% \paragraph{PCK}
\textit{Percentage of correct keypoints (PCK)} metric calculates the percentage of correctly predicted keypoints, where a predicted keypoint is thought correct if its distance to its target is smaller than a threshold $\delta$:
\begin{equation}\label{eq:PCK}
    % \Scale[0.9]{   
    L_\text{PCK} = \frac{1}{T\times J}\sum_{t=1}^T \sum_{j=1}^J{\mathbf{1}\big[\Vert p^j_t - \hat{p}^j_t\Vert_2<\delta\big]},
    % },
\end{equation}
where $\mathbf{1}$ is the indicator function and $p^j_t$ indicates joint $j$'s position of frame $t$.
The threshold $\delta$ is set to 0.2 in our experiments as in~\cite{speech2gesture}.

\vspace{2mm}
% \paragraph{STFT}
\textit{Short-term Fourier transform (STFT)} is used to compare the motion dynamics between the generated motion and the groundtruth (Eq.~\ref{eq:ssim}, Sec.~\ref{sec:ssim}).
% To compare the motion dynamic between generated motion and the groundtruth, STFT is also introduced as a metric.
In our experiments, the window size and stride size and FFT size of the STFT are set as 32 and 8 respectively (\textit{i.e.} 1.06/0.265 sec for 30 FPS motion).
This metric is similar with PSKL~\cite{motioninpainting} when the window covers the entire motion sequence.
% is same with the length of generated motion.

\vspace{2mm}
\textit{Structural similarity index measure (SSIM)} measures the dynamics and the structure similarity at same time (Eq.~\ref{eq:ssim}, Sec.~\ref{sec:ssim}).

\vspace{2mm}
\textit{Learned perceptual image patch similarity (LPIPS)}~\cite{lpips} measures the similarity between the generated motion and the groundtruth in a high-level feature space that considers both spatial and temporal information (Eq.~\ref{eq:lpips}, Sec.~\ref{sec:lpips}).
The network used to extract this feature is trained with motion reconstruction task on Trinity dataset.
We use the average pooled feature across time axis to represent a motion clip in LPIPS and FID calculation.

\vspace{2mm}
\textit{Fréchet inception distance (FID)~\cite{fid}} measures the similarity between a collection of generated motions and a collection of groundtruth motions by calculating the Fréchet distance on the extracted deep motion features.
The deep features are extracted using the same network as LPIPS but only the output of the $4^{th}$ residual block is used, which is then averaged across the temporal dimension.
The final deep motion features is a 128 dimension vector.
The Fréchet distance~\cite{fid} is calculated by:
\begin{equation}
    L_\text{FD} = \Vert \mu_1 - \mu_2 \Vert^2 + \operatorname{Tr}(\Sigma_1+\Sigma_2-2(\Sigma_1\Sigma_2)^{\frac{1}{2}}),
\end{equation}
where $\mu$ and $\Sigma$ is the mean and the variance of the motion features.

\subsubsection{Diversity metric} \label{sec:metric:diversity}

Diversity measures how many different poses/motions have been generated within a long motion.
People can easily notice the motion is not performed by a real person or get bored if only static motions or repeated motions can be generated.
For example, RNN-based methods easily get stuck into some undesired static motion as the generated motion becomes longer and longer. 
% And static motions, which are undesired apparently, should get low diversity scores.

We first split the generated motions into equal-lengthed non-overlapping motion clips (50 frames per clip in our experiments) and we calculate diversity as the averaged $L_1$ distance of these short motion clips.
Formally, it is defined as:
\begin{equation}\label{eq:diversity}
% \Scale[0.9]{ 
    L_\text{Diversity} = \frac{1}{N \times \ceil{N/2}} \sum_{a_{1}=1}^N \sum_{a_{2}=a_{1}+1}^N \Vert \hat{M}_{a_1} - \hat{M}_{a_2}\Vert_1 
% },
\end{equation}
where the $\hat{M}_{a_1}$ and $\hat{M}_{a_2}$ represent clips from the same motion sequence, $N$ represents the count of the motion clips, which is $\frac{T}{50}$ in our experiments.
Please note that jitter motion and invalid poses can also result in high diversity score, which means higher diversity is preferred only if the generated motion is natural.
% So higher diversity is preferred only if the generated motion is natural.

\subsubsection{Multimodality} \label{sec:metric:multimodality}
Multimodality measures how many different motions could be sampled (through multiple runs) for a given audio clip.
This is important due to the inherent one-to-many mapping between audio and motion. 
Note that multimodality calculates motion difference across different motions
while diversity calculates the difference among short motion clips within the same long motion.
We measure the multimodality by generating motions for an audio $N$ times, which is 20 in our experiments, and then calculate the average $L_1$ distance of the motions.
\begin{equation}
    % \Scale[0.9]{
    L_\text{Multimodality} = \frac{1}{N \times \ceil{N/2}} \sum_{a=1}^N \sum_{b=a+1}^N\Vert\hat{M}_a-\hat{M}_b\Vert_1
    % },
\end{equation}
where the $\hat{M}_a$ and $\hat{M}_b$ represent
sampled motions generated through different runs for the same audio input.
Similar to diversity, invalid motion will also result in abnormally high multimodality score.

\subsubsection{User studies} \label{sec:metric:user_study}

% Since all the existing quantitative metrics can measure one aspect, 
% To evaluate the results qualitatively, 
We conduct user studies to analyze the visual quality of the generated motions since every metric measures only one aspect and they do not align well enough with human evaluations.
Two use studies are conducted to compare 1) different methods and 2) to compare the newly introduced motion losses.
For the first questionnaire, it contains four 20-second long videos. The motion clips shown in one video is generated by various methods from the same audio clip.
The participants are asked to rank the motion clips from the following three aspects respectively: 
\begin{enumerate}
\item Realism: which one is more realistic?
\item Diversity: which motion has more details?
\item Matching degree: which motion matches the audio better?
\end{enumerate}
The results of the questionnaires are shown in \reffig{fig:user_study}.
We show the count of different ranking in the figure. The average score of different metrics for each algorithm is listed after the corresponding bar.
The scores assigned to each ratings are \{5,4,3,2,1\} for \{best, fine, not bad, bad, worst\} respectively.
% \dknote{any interesting findings? \textit{e.g.} which metric aligns well with the user study.}

% \TODO{update, the details of two user studies are different}
% \lj{
For the second questionnaire, it contains 12 videos ranging in length from 10 seconds to 30 seconds.
The motion clips shown in one video is generated by different methods (i.e. adding different motion losses from STFT, SSIM, or LPIPS) given the same audio clip.
The participants are asked to rank the motion clips according the realism and diversity of the motion.
The results of the questionnaires are shown in \reffig{fig:user_study_abla}.
We show the count of different ranking in the figure. The average score of different metrics for each algorithm is listed after the corresponding bar.
The scores assigned to each ratings are \{4,3,2,1\} for \{best, fine, not bad, worst\} respectively.
% }

% ********* ********* *********
%\begin{table}[!t]
%% increase table row spacing, adjust to taste
%\renewcommand{\arraystretch}{1.3}
% if using array.sty, it might be a good idea to tweak the value of
% \extrarowheight as needed to properly center the text within the cells
%\caption{An Example of a Table}
%\label{table_example}
%\centering
%% Some packages, such as MDW tools, offer better commands for making tables
%% than the plain LaTeX2e tabular which is used here.
%\begin{tabular}{|c||c|}
%\hline
%One & Two\\
%\hline
%Three & Four\\
%\hline
%\end{tabular}
%\end{table}

% \begin{table*}[!t]
% \caption{The necessary of the Motion Dictionary. }% \dknote{not very important if the Dance experiments go well.}
% \label{tab:motion_dict}
% {\def\arraystretch{1}\tabcolsep=3.5em
% \begin{tabular}{l|r@{\hspace{0.1cm}}lr@{\hspace{0.1cm}}lr@{\hspace{0.1cm}}lc}
% \toprule[1pt]
%      Motion Dict & \multicolumn{2}{c}{$L_1$ $\downarrow$}  & \multicolumn{2}{c}{ PCK $\uparrow$ } &\multicolumn{2}{c}{ Diversity $\uparrow$} & Multimodality $\uparrow$ \\
% \midrule[1pt]
%     w/o Motion Dict\\
%     Training data (VAE) &  \\
%     Learned Motion Dict (VQ-VAE) &  \\
% \bottomrule[1pt]
% \end{tabular}
% }
% \end{table*}

% ********* ********* *********
\subsection{Implementation details}
\label{sec:implementation}
\subsubsection{Data processing} \label{sec:data_proc}
We detail the data processing of Trinity dataset and S2G-Ellen dataset here.

\noindent\textbf{Trinity dataset.}
The audio data are resampled to 16kHz for extracting log-mel spectrogram~\cite{logmel} feature using librosa~\cite{librosa}.
More concretely, the hop size is set to $ \text{SR} / \text{FR} $ where $\text{SR}$ is the sample rate of the audio and FR is the frame rate of the motion so that the resulting audio feature have the same length as the input motion.
In our case, the resulting hop size is 533 since $\text{SR}$ is 16000 and $\text{FR}$ is 30.
The dimension of the log-mel spectrogram is 64.

\par
The motion data are downsampled to 30 FPS and then retargeted to the SMPL-X~\cite{SMPL-X} model.
SMPL-X is an expressive articulated human model consisting of 54 joints (21 body joints, 30 hand joints, 3 face joints, respectively) , which has been widely used in 3D pose estimation and prediction~\cite{hmr,SMPL-X,phd,spin}.
The joint rotation is in 6D rotation representation~\cite{6d_repr} in our experiments, which is a smooth representation and could help the model approximate the target easier. 
Note that the finger motions are removed due to unignorable noise.

\noindent\textbf{S2G-Ellen dataset.} 
Following~\cite{speech2gesture}, the data are split into 64-frame long clips (4.2 seconds). 
Audio features are extracted in the same way as the Trinity dataset.
%The 64-dimension log-mel spectrogram is extracted with librosa.
The 2D body joints are represented in a local coordinate frame relative to its root (\textit{i.e.} origin is the root joint location) on the image plane.
% Namely, the origin of the coordinate is the root joint.

\subsubsection{Network}
Most of our experiments take the TCN as the backbone of the encoders unless stated otherwise.
Every encoder and decoder consists of 5 residual blocks~\cite{resnet}, each containing several 1D convolution layers with ReLU non-linearity~\cite{relu}. 
The residual block is similar to~\cite{TCN} except several modifications. 
Specifically, we use normal symmetric 1D convolutions that see both the history and the future instead of casual convolutions that see only the history.
% the casual convolutions whose kernels see only the history are replaced with normal symmetric 1D convolutions seeing both the history and the future.
% 
Similar to the bottleneck setting in the MLP-based autoencoders, the channel numbers of convolution layers are set to \{128, 128, 96, 96, 64\} for the audio encoder, \{256, 256, 128, 128, 64\} for the motion encoder, and \{64, 128, 128, 256, 256\} for the motion decoder.
The kernel size (on time axis) of the convolution layers in backbones is 3.
Two parallel $1\times1$ convolution layers are appended to the encoders to predict the mean and variance of the shared latent features. 
For non-variational case, only one $1\times1$ convolution layer is needed.
% \dknote{implemented with conv layers? Otherwise, how can you get a sequence of latent codes?}.
% 
The motion specific feature is sampled from Gaussian distribution, whose mean and variance is predicted from two parallel linear layers, respectively.
Both the shared code and motion-specific code are set to 16 dimensions. 
% \dknote{also add kernel size}
% \dknote{need more configuration details, \textit{e.g.} channels}
% More details are included in the supplementary.

\subsubsection{Training}
At the training stage, we randomly crop a 4.2-second segment of the audio and motion data, which is 64 frames for the S2G dataset (15 FPS) and 128 frames for the Trinity dataset (30 FPS).
The model weights are initialized with the Xavier method~\cite{glorot2010understanding} and trained 180K steps using the Adam~\cite{adam} optimizer. 
The batch size is 32 and the learning rate is $10^{-4}$.
The $\lambda_{\text{rot}}$, $\lambda_{\text{pos}}$, $\lambda_{\text{speed}}$ are set as ${1,1,5}$ respectively,
and $\rho$ is set as $0.2$ cm in our experiments.
Our model is implemented with PyTorch~\cite{pytorch}.

\section{Results and discussions}

\subsection{Comparison with state-of-the-art methods} \label{sec:sota}
\input{tab_tex/sota}
\begin{figure}[t]
\centering
\includegraphics[width=\columnwidth,trim={0.2cm 0.8cm 0.6cm 1cm},clip]{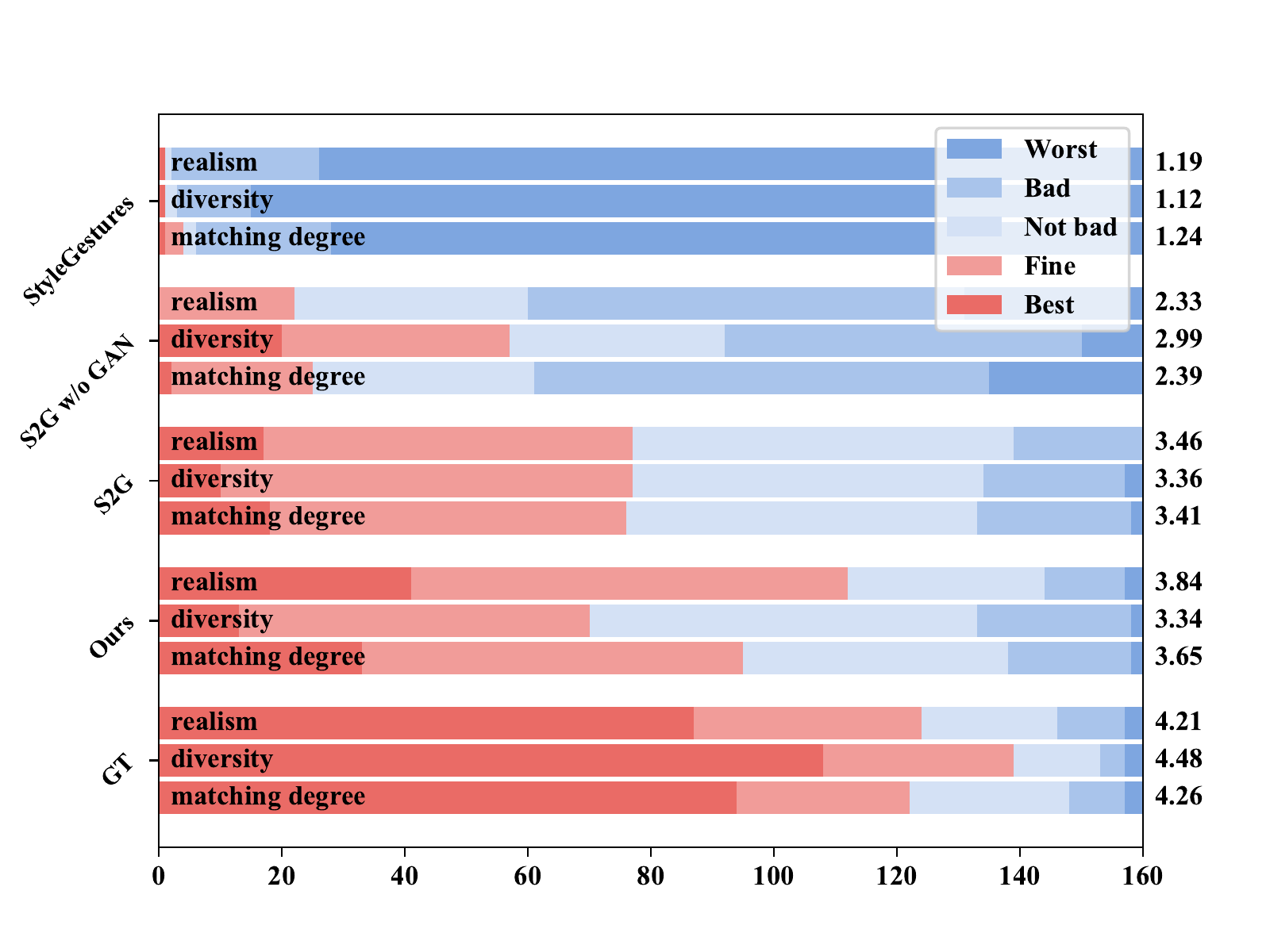}
\DeclareGraphicsExtensions.
\vspace{-8mm}
\caption{
User study results to compare our method against previous state-of-the-art methods. 
The horizontal axis represents the number of samples rated by the participants. 
In total, 160 comparisons have been rated (40 participants, 4 comparisons each questionnaire).
Colored bars in different lengths indicate the counts of their corresponding rankings.
The average score (higher is better) for each method is listed on the right.
``S2G'' is short for Speech2Gesture~\cite{speech2gesture}. 
User study videos for comparison are provided in the supplementary.
Also see Tab.~\ref{tab:sota_result} for quantitative evaluations.
}
\label{fig:user_study}
\end{figure}

\begin{figure*}[t]
\centering
\includegraphics[width=\textwidth,trim={0.cm 0cm 0.cm 1.cm},clip]{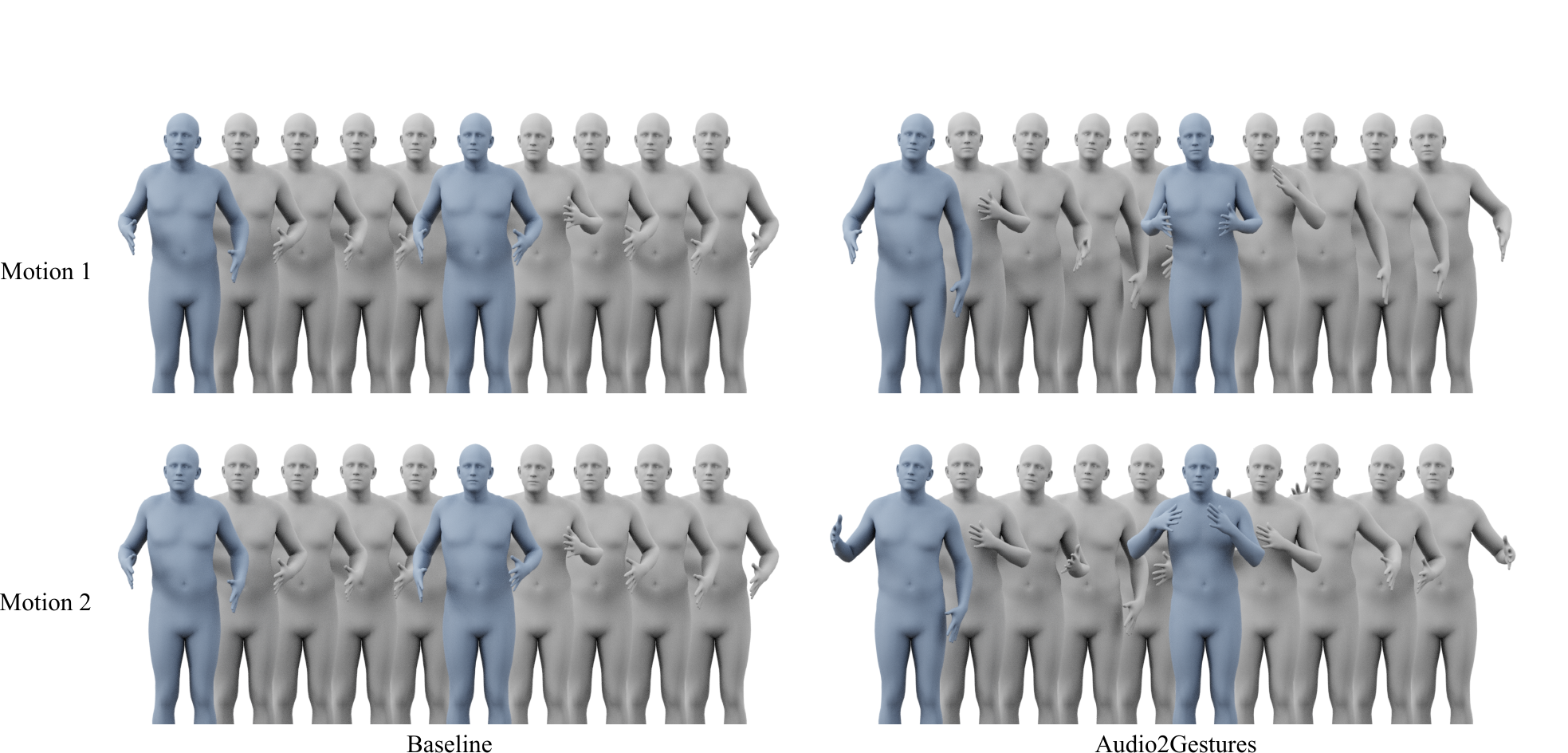}
\vspace{-8mm}
\caption{Two examples (rows) of the generated multimodal motions given the same audio input using baseline method and our full method in Tab.~\ref{tab:ablation_components}. 
% \dknote{baseline refers to what setting?}
Two different motions generated by the baseline method have minimal difference since its 
 multimodality score is only 0.41.
In contrast, different motions predicted by our full model have certain similarity (\textit{e.g.} lifting hands from waist) but not the same.
% The predicted motions are similar (\textit{e.g.} lifting hands from waist) and not the same.
% which is mainly caused by the training dataset is relative small and not all the audio feature maps to multi-pose, thus the motions are mainly controlled by the audio signal.
% Some motion, shown in blue avatar, are very different, which indicates that the model could model the audio-motion one-to-many mapping.
}
% \todo{need update the figure and the caption}
\label{fig:result}
\end{figure*}

We compare our method with two recent representative state-of-the-art methods, including one LSTM-based method named StyleGestures~\cite{stylegestures} and one CNN-based method named Speech2Gesture~\cite{speech2gesture} on Trinity dataset.
StyleGestures adapts normalizing flows~\cite{normalizing_flow,glow,moglow} to speech-driven gesture synthesis.
We train StyleGestures using the code released by the authors. 
The training data of the StyleGestures are processed in the same way as the authors indicate\footnote{The motions generated by StyleGestures are 20 FPS and have a different skeleton from our method. We upsample the predicted motion to 30 FPS and retarget it to SMPL-X skeleton with MotionBuilder.}.
Speech2Gesture, originally designed to map speech to 2D human keypoints, consists of an audio encoder and a motion decoder. 
Its final output layer has been adjusted to predict 3D joint rotations and is trained with the same losses as our method. 

Quantitative experimental results are listed in \reftab{tab:sota_result} and user study results in \reffig{fig:user_study}.
The results on Trinity show that our method outperforms previous state-of-the-art algorithms on the realism (see \reffig{fig:user_study}) and diversity metrics (see \reftab{tab:sota_result} and \reffig{fig:user_study}), 
demonstrating that it is beneficial to explicitly model the one-to-many mapping between audio and motion in the network structure. 
% \dknote{This is for trinity only. Discuss Ellen results.}

While StyleGestures supports generating different motions for the same audio via sampling, the quality of its generated motions is not very appealing (see \reffig{fig:user_study}).
% \dk{see \reffig{fig:user_study}?} \dknote{need support for this statement, since ``quality'' is not a specific word}.
%
Also, its diversity score is the lowest, because LSTM output easily gets stuck into some fixed poses, resulting in long static motion afterwards.
The algorithm is not good at generating long-term sequences due to the error accumulation problem of the LSTM.
% 
% The authors test their algorithm on 400 frames (13 seconds) length sequences.
The LSTM-based StyleGestures has only been tested on up to 400-frame (13-second) long sequences originally.
However, obviously deteriorated motions are generated when evaluating their algorithm to generate 5000-frame (166-second) long motions.

As for Speech2Gesture, the generated motions show similar realism with ours but obtain lower diversity score (\reftab{tab:sota_result}) than our method.
But Speech2Gesture does not support generating multimodal motions given the same audio input.
Also note that Speech2Gesture with GAN generates many invalid poses and gets the worst performance.
We have trained the model several times changing the learning rate range from 0.0001 to 0.01, and report the best performance here.
The bad performance may be caused by the unstable of the training process of the generative adversarial network.

% \lj{
% Considering that Speech2Gesture~\cite{speech2gesture} is originally designed for generating 2D gestures, 
We also compare our method for 2D co-speech joint generation task on S2G-Ellen data~\cite{speech2gesture} with Speech2Gesture~\cite{speech2gesture}, which is originally designed for this 2D generation task.
It can be seen that if GAN loss~\cite{gan} is not used, the Speech2Gesture~\cite{speech2gesture} model gets the best $L_1$ and PCK score but the worst Diversity performance, which is 0.61.
We can clearly see the trade-off between better similarity metrics (\textit{i.e.} $L_1$, PCK) and the diversity metrics (\textit{i.e.} diversity, multimodality).
``S2G w/o GAN'' achieves best similarity scores and worst diversity scores while ``S2G'' the opposite.
In contrast, our method strikes a good balance between them.
% 
% The Diversity performance of Speech2Gesture could be improved to 0.89 after introducing the GAN loss~\cite{gan}, but the $L_1$ and PCK metric drops.
% All the metrics of our method is between the Speech2Gesture and Speech2Gesture w/o GAN.
% }

% \begin{figure}[t]
% \centering
% \includegraphics[width=\columnwidth,trim={0cm 0.8cm 0.5cm 1cm},clip]{img/user_study2.pdf}
% \vspace{-8mm}
% \caption{
% User study results to compare different variants of the full method. 
% % Results of the user study where we compare our method with its variants.
% The horizontal axis represents the total number of samples rated by the participants.
% In total, 100 comparisons have been rated (25 participants, 4 comparisons each questionnaire).
% The lengths of different colored bars indicate the counts of their corresponding ratings.
% The average score for each row is listed on the right.
% Colored bars in different lengths indicate the counts of their corresponding rankings.
% User study videos for comparison are provided in the supplementary.
% Also see Tab.~\ref{tab:ablation_components} for quantitative evaluations.
% \dknote{This figure is not mentioned in main text.}
% % \dknote{trim white spaces}
% }
% \label{fig:user_study_abla}
% \end{figure}

\subsection{Ablation study} \label{sec:ablation}

\input{tab_tex/ablation_components}

\input{tab_tex/ablation_relax_motion_loss}

In this section, we first verify the effectiveness of the proposed major modules in Sec.~\ref{sec:ablation:modules}.
Then we test the influence of different threshold values in the relaxed motion loss in Sec.~\ref{sec:ablation:th_for_relax}.
In Sec.~\ref{sec:result:new_motion_losses}, we study whether or not the introduced new motion losses can improve the quality of the generated motions.
Finally, we conduct experiments to show that the proposed split formulation are applicable for different backbones in Sec.~\ref{sec:ablation:backbones}.

\subsubsection{Modules} \label{sec:ablation:modules}

To gain more insights into the proposed components of our model, we test some variants of our model on the 3D Trinity dataset (\reftab{tab:ablation_components}).
We run every variant 20 times and report both {\it average} performance and the {\it best} performance (listed in parentheses).
% to avoid the influence of randomness.
Note that the variation of our model comes from two different parts, the randomness introduced by the variational autoencoder and by the motion-specific feature sampling.
\input{tab_tex/ablation_loss}

\par
We start with a ``baseline'' model, which excludes the mapping net and the split code. 
It is trained with only the motion reconstruction losses (Eq.~\ref{eq:mr}) and shared code constraint (Eq.~\ref{eq:sca}).
The average scores of Pos. $L_1$, PCK, and diversity have little difference with their best scores,
% ``avg $L_1$'', ``avg PCK'' and ``avg Diversity'' of the model equal to the best scores ``min $L_1$'', ``max PCK'' and ``max Diversity'' on $L_1$, 
which indicates that the randomness of the VAE model alone have very little influence on generating multimodal motions.

\par
The next setting is ``+split'', which splits the output of the motion encoder into shared and motion-specific codes and introduces the relaxed motion loss (Eq. \ref{eq:rmr}). 
This modification greatly improves the multimodality score by explicitly handling the one-to-many mapping (from 0.41 to 5.95), but it harms the similarity metrics (see ``Pos. $L_1$'', ``Speed'', ``Acc.'', ``PCK'' and LPIPS).
This huge performance drop might be caused by the misalignment between the sampled signal and the motion-specific feature.
% 
% \lj{
For example, we find a big difference in the statistics (\textit{i.e.} the mean and variance) of the delta of motion-specific feature (between two consecutive frames) between the training samples and the randomly sampled one.
% }

\par
Thus, training set statistics and a StyleGAN fashion mapping network (``+mapping net'') are introduced to address the potential misalignment between the sampled signal and the real motion-specific feature automatically.
%
% Although the multimodality drops compare to ``+split'', this modification helps to improve other metrics of the generated motions a lot.
% Note that a higher multimodality score only makes sense when the generated motions is natural, as described in \refsec{sec:eval}. \dknote{Why do we need to stress it here.}
The ``+mapping net'' model also outperforms the baseline model in the Pos., Speed, Acc., STFT 
, FID and Diversity metrics and gets a similar PCK metrics, but the model is able to generate multimodal motions.
%
% We notice that the diversity of the motions generated by ``+mapping net'' model is not as good as the baseline model.
% The $L_1$ score of the ``+mapping net'' model is also worse than the baseline model, which may be due to the users prefer the motions with more dynamics.
%
% We think the problem may be caused by the mode collapse problem suffered by many generative methods.

Because generative models often suffer from mode collapse problem, two simple yet effective losses -- the bicycle constraint (Eq.~\ref{eq:cr}) and a diversity loss (Eq.~\ref{eq:ds}) -- are introduced to alleviate the problem.
Bicycle constraint improves the multimodality of the motions from 3.59 to 3.99.  
The diversity loss further improves the multimodality to 4.10 but has little influence on the similarity metrics.\ref{tab:ablation_components}
%
% The final model outperforms the baseline model in all quantitative indicators, which shows that the audio-motion mapping could be better modeled by explicit modeling the one-to-many correlation.
\subsubsection{Thresholds in relaxed motion loss}  \label{sec:ablation:th_for_relax}

In \reftab{tab:relaxed_motion_loss}, we show the influence of the hyper-parameter $\rho$ of the relaxed motion loss.
When the threshold $\rho$ is set to 0, the relaxed motion loss becomes the common $L_1$ loss, and the algorithm gets the best $L_1$, PCK and LPIPS performance (7.42/0.83/44.2).
As the $\rho$ increases, the sampled motions are less penalized if they are not very similar to their corresponding groundtruth, resulting in gradually decreased performance with respect to similarity metrics.
For example, the $L_1$ (LPIPS) error increases to 8.04 (0.79) and the PCK drops to 0.79 when $\rho$ is set to 0.2.
However, worse $L_1$, PCK, and LPIPS performance do not necessarily mean the quality of the generated motion is not good due to the inherent weak and multimodal correlation between audio and motion especially for long-term motion synthesize.
The FID is 1.66 when the $\rho=0.2$ and is better than $\rho=0$ whose FID is 1.81.
What is more, the multimodality of the generated motion gradually getting better from 3.59 to 4.54, which indicates that more diverse motions could be generated given the same input audio (\textit{i.e.} better multimodality).
% the difference of the motions that generated conditioned on the same audio gets larger.
% \dknote{need mention what have been changed for experiments in Tables 1 and Tables 2/3/4.}
% \lj{
% The results of Audio2Gestures in \reftab{tab:sota_result} and \reftab{tab:ablation_components} are little different, which is mainly caused by the randomness of the algorithm.
% }

\subsubsection{Additional motion losses}  \label{sec:result:new_motion_losses}

We study the influence of three new motion losses (STFT, SSIM, and LPIPS) in Tab.~\ref{tab:ablation_loss}.
The basic motion reconstruction loss (Eq.~\ref{eq:mr}) is always present.

These losses, in complement to commonly used point-wise losses in Eq.~\ref{eq:mr}, measure the similarity of two motions with the consideration of spatial and/or temporal context.

They could also be viewed as relaxed motion losses from the perspective that they focus more on measuring the local structures aside from values point-by-point (\textit{e.g.} $L_1$ distance), resulting in a trade-off between accuracy and quality (see the similarity metrics and other metrics in Tab.~\ref{tab:ablation_loss}).

How to quantitatively measuring motion quality is an underdeveloped problem. 
% \lj{
%     In this section, we take a step toward this direction by comparing the scores of different metrics with the user study result.
The frame level metrics (Pos. $L_1$, PCK) are widely used metrics, which are standard metrics of short-term motion prediction.
However, the metrics are hard to reflect the performance of the long-term motion generation tasks, because of the audio-motion one-to-many mapping. 
For example, expressive ourselves with left hand or right hand have little difference in most occasions but have a large difference in Pos. $L_1$ score.
% Another limitation of the metrics is they do not consider the complicated motion dynamics. Mean pose and jitter motion could both get a good $L_1$ score.
% loss is often used at training the network, thus the metric could evaluate  how well the motion model is trained in some degree.

Another limitation is that the motion dynamics (\textit{e.g.} smoothness) is one crucial factor that affects the user study results.
% \dk{
However, we find the most commonly tested metrics (\textit{i.e.} Pos. $L_1$, PCK, and FID) can hardly tell whether the motion is smooth or jerky, and thus introducing speed, acceleration (Acc.), STFT, and SSIM as complements.
% We notice that the smoothness of the generated motion is one of the most important factor that affects the results of user study.
% To analyze the alignment of the qualitative metrics with the quantitative metric, 
To show their effectiveness, we add two levels of Gaussian noise to the Euler angles on every joint of the GT motion.
Their results are listed in \reftab{tab:ablation_loss} for reference.
% to the groundtruth motion and calculate the score of the qualitative metrics (The last row of \reftab{tab:ablation_loss}).
Although the noisy motions jitter very much, Pos. $L_1$, PCK, LPIPS, and FID are very insensitive to these obvious artifacts because Pos. $L_1$, PCK completely do not consider temporal information and LPIPS and FID measure distance in high-level feature space, which might have ignored this low-level difference.
In contrast, the Speed $L_1$, Acc. $L_1$, STFT, and SSIM can evaluate smoothness faithfully.
% }
% thus a good metric is expected to gave a worse score.
% From the results we can notice that Speed, Acc, STFT and SSIM works well at evaluating the smoothness of the motion.
% The jitter of motion have little influence on LPIPS and FID metrics, which may caused by the motion feature extracted by our Motion Autoencoder is not high-level enough. 
% }

We notice that the motions predicted by the network trained with STFT loss usually contain more subtle motion details and they perform better on Speed, Acc., PCK, STFT, and SSIM metrics (from \{0.6678/0.2432/0.8130/1.13/0.8594\} to \{0.6565/0.2255/0.8134/1.02/0.8621\}).
We conduct user study in Sec.~\ref{sec:metric:user_study} for qualitative evaluation.
According to the feedback, using STFT as training loss indeed brings noticeable improvement from the perspective of motion quality.

\begin{figure}[t]
\centering
    \includegraphics[width=\columnwidth,trim={0cm 0.8cm 0.4cm 1cm},clip]{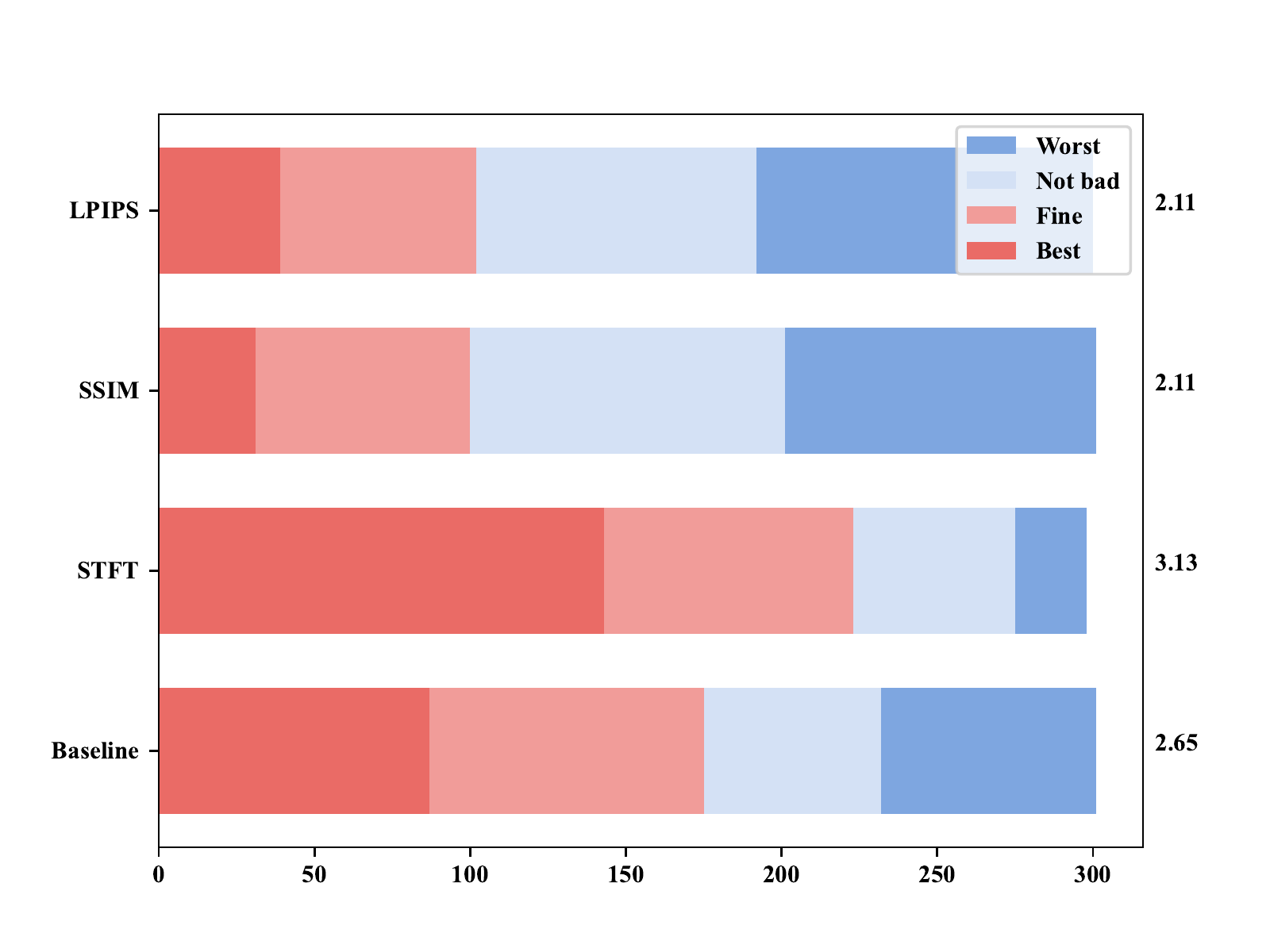}
\vspace{-8mm}
\caption{
User study results to compare the newly introduced motion losses. This user study specifically focus on the naturalness of the generated motion (i.e. no dedicated diversity and matching degree questions.)
% Results of the user study where we compare our method with its variants.
The horizontal axis represents the total number of samples rated by the participants.
In total, 300 comparisons have been rated (25 participants, 12 comparisons each questionnaire).
% The lengths of different colored bars indicate the counts of their corresponding ratings.
Colored bars in different lengths indicate the counts of their corresponding rankings.
The average score for each row is listed on the right.
User study videos for comparison are provided in the supplementary.
Also see Tab.~\ref{tab:ablation_loss} for quantitative evaluations.
% \dknote{trim white spaces}
}
\label{fig:user_study_abla}
\end{figure}

% \lj{
When trained with SSIM loss, although the Pos., Speed, Acc., PCK, STFT, and SSIM metrics improve a lot (\{7.8623/0.6384/0.2211/0.8167/1.10/0.8647\}),
no noticeable improvement has been observed according to the user study (\reffig{fig:user_study_abla}).

LPIPS evaluates the similarity of the generated motion with the target motion in learned motion feature space instead of directly comparing the low level $L_1$ distance of joint position.
Although the metric often corresponding well with the positional $L_1$ metric in our experiments, it improves the multimodality of the generated motion a lot after adopted as a training loss, which raise to 4.68 from 3.57.
Although the averaged Pos. $L_1$ and LPIPS drops to 8.33 and 48.3 from 7.84 and 47.0, the best performance of the metrics improved to 7.63 and 43.5.
The FID and speed $L_1$ also better than the naive Audio2Gestures model.

% \paragraph{The influence of the relaxed motion loss}

\subsubsection{Backbones}  \label{sec:ablation:backbones}
\input{tab_tex/backbone}
\input{tab_tex/dct}

At the core of our Audio2Gestures is the split latent space and the strategies to better learn the latent codes.
It does not require particular network architectures, so we study whether or not this formulation is applicable for other backbones, including GRU and Transformer, in this section.
We replace the TCN backbone to extract latent codes with GRU-based~\cite{gru} and Transformer-based~\cite{transformer} network and their results are listed in \reftab{tab:backbone_ablation_study}.
Notice that the decoder to transform the latent codes to final motion remains unchanged in these experiments.

To make fairer comparisons, we use the same number of layers as the original TCN backbone.
Concretely, the GRU backbone consists of 5 bidirectional GRU layers, each with 32 hidden neurons.
The transformer backbone consists of 5 transformer encoder layers. 
The dimensions of the Q, K, and V of the self-attention are all set to 16, where Q, K, and V use separate learnable transformations and only one head is used.
The following MLP of each transformer layer contains two FC layers, projecting the 16-dim feature to 256 and back to 16 (both using ReLU activation) for non-linear feature transformation.
Similar to DanceRevolution~\cite{dance_revolution}, we apply a sliding window mask at the transformer encoders so that the gestures do not depend on the information that far from the current speech signal.
The window size is set to 32, which is about 1.06 seconds.
The sequence length at the training stage is set as 512 (\textit{i.e.} 17.06 seconds).

\par
They can all generate smooth motions through our visual inspection.
The quantitative results are listed in Tab.~\ref{tab:backbone_ablation_study}.
Using GRU as the backbone gives a similar similarity metric, where the Pos., PCK, STFT, and SSIM dropped but the Speed and Acc. improved.
The transformer backbone gets a better similarity metric (\textit{e.g.} Pos. $L_1$, Speed $L_1$, Acc. $L_1$ and PCK) but worse motion dynamics (\textit{e.g.} STFT, Diversity, also see the supplemented video). 
The multimodality of the Transformer backbone is also worse than the TCN and GRU backbone.
We notice the candy-wrapper effect of the elbow joints occasionally due to bad rotation configuration for the Transformer backbone.

\subsection{Latent DCT space}

% Inspired by MOJO~\cite{mojo}, which models human motion in the discrete cosine transform (DCT) space to predicts different future motion conditioned on previous motion by sampling diverse high frequency bands, our method could modeling the features within DCT space, which enables us control the frequency of the generated motion by editing the high-frequency bands of the audio-motion shared feature or generate diverse motion by sampling diverse motion-specific feature within DCT motion-specific space.

Similar to MOJO~\cite{mojo}, our latent codes could also be modeled in discrete cosine transform (DCT) space to support sampling per frequency band.
% 
% \dknote{This is a result section. Only need to briefly mention the benefit. Focus on implementation details and result discussion.}

For example, we can insert a DCT right before the two FC layers predicting the mean and standard deviation (\textit{i.e.} the latent space of a VAE) and insert an IDCT after them.
With this modification, a sequence of time domain features is transformed to DCT coefficients (128 points in our experiments) and then back to the time domain, enabling frequency domain controls.
In our experiments, we test inserting DCT and IDCT for both the shared feature and the motion-specific feature. 
But only the motion-specific branch adopts VAE while the shared feature branch adopts AE (\textit{i.e.} only one FC layer)
since we do not have paired motion and have to sample motion-specific code during inference.
Note that the encoders require fixed length input due to the extra length constraint posed by the DCT.
So we split long test audio into fixed-length clips and extract all their latent features (time domain feature after IDCT) using the encoders in advance for this DCT variant.
Then all the features are concatenated and then decoded to motions by the following TCN in a fully convolutional fashion as before.

\par
In \reftab{tab:feat_dct}, we show the quantitative results of motions generated by editing the features in the DCT domain.
The \textit{Audio2Gestures w/ DCT} achieves similar performance with the original Audio2Gestures model.
Then we test various frequency manipulation in the feature domain. 
Note that all the generated motions are still smooth (\textit{i.e.} do not jitter) and valid (\textit{i.e.} no invalid poses).

For audio-motion \textit{shared code} $S_A$, 
all metrics gradually get worse as we drop (\textit{i.e.} set to 0) the highest frequency bands of the audio features (see \reftab{tab:feat_dct}, middle) and only keeps the lowest \{100, 50, 10\} frequency bands, which indicates that the motions generated by Audio2Gestures w/ DCT are correlated with the input audio.
We observe the generated motion is very slow after dropping too many high-frequency components, which, to some extent, shows the dynamic of the generated motion is well correlated to its audio control.
% 
% It can be also noticed that the Diversity and the STFT metric gradually get worse, which indicates that the frequency of the generated motion is correlated with the frequency of the audio feature. 
% \dk{Note that we did not use STFT as loss during training in these experiments.}

For \textit{motion-specific code} $I_R$, only a few low-frequency bands (around 10) have an obvious influence on the final motion (see \reftab{tab:feat_dct}, bottom).
This might be the result of that we encourage the audio-motion shared latent code to contain more information so that the input audio can control the generated motion effectively (see Sec.~\ref{sec:latent:align}).
% , which makes sense since we expect the shared feature conveys major information (\textit{e.g.} meaning, beat) while the motion-specific feature conveys less important variations. \dknote{update later}
% \dknote{check. It is contradictory to our application.}
% the generated motions do not change as obvious as dropping 
% if we gradually drops the highest frequency of the sampled motion-specific feature ($I_R$), the benchmarks drops little even if we only keep the lowest frequency band.
% 
But when we drop all the frequency bands ($I_R=0$), all the metrics drop a lot, which means the motion specific-code is not degenerate and plays an important part in our formulation.

\subsection{Application}

\begin{figure*}[t]
\centering
\includegraphics[width=\textwidth,trim={0cm 0.2cm 0cm 0cm},clip]{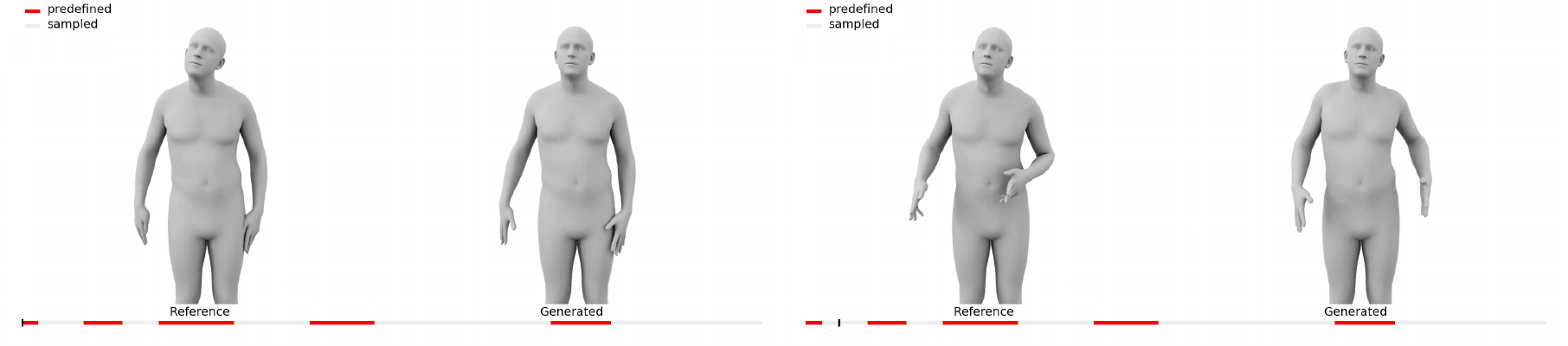}
\vspace{-4mm}
\caption{Left: The generated motion is different from the reference motion because the motion-specific code is randomly sampled. Right: The generated motion is similar to the reference motion because the motion-specific code is extracted from the reference motion using the motion encoder.}
\label{fig:application_demo}
\end{figure*}

We notice that motion-specific code extracted from a motion strongly controls the final motion output.
To be specific, the synthesized motion is almost the same as the original motion used to extract this motion-specific code.
This feature is perfect for a type of motion synthesis application where pre-defined motions are provided on the timeline as constraints.
For example, if there is a $n$-frame long motion clip that we want the avatar to perform from frame $t$ to $t+n$.
We could extract its motion-specific code $I_M$ with the motion encoder and directly replace the sampled motion-specific code $I_R$ from $t$ to $t+n$.
Our model could generate a smooth motion from the edited motion-specific code. 
Please refer to our project page for the demonstration.

%%%%%%%%%%%%%%%%%%%%
%%%% Conclusion %%%%
%%%%%%%%%%%%%%%%%%%%
\section{Conclusion}
In this paper, we explicitly model the one-to-many mapping by splitting the latent code into shared code and motion-specific code. 
This simple solution with our customized training strategy effectively improves the similarity, diversity, and multimodality of the generated motion.
We also demonstrate an application that the model could insert a specific motion into the generated motion by editing the motion-specific code, with smooth and realistic transitions.
Despite the model could generate multimodal motions and provide users the ability to control the output motion, there exist some limitations. For example, the generated motion is not very related to what the person says.
Future work could be improving the meaning of the generated motion by incorporating word embedding as an additional condition.

% \clearpage
% or


% \appendices

% % you can choose not to have a title for an appendix
% % if you want by leaving the argument blank
% \section{}

% \section*{Acknowledgment}


\ifCLASSOPTIONcaptionsoff
  \newpage
\fi



\bibliographystyle{IEEEtran}
\bibliography{IEEEabrv,lijing}
\end{document}

%% file: img/fig_network.tex
\begin{figure*}[!t]
\centering
\begin{overpic}[trim=3.5cm 6.6cm 5cm 3.2cm,clip,width=1\linewidth,grid=false]{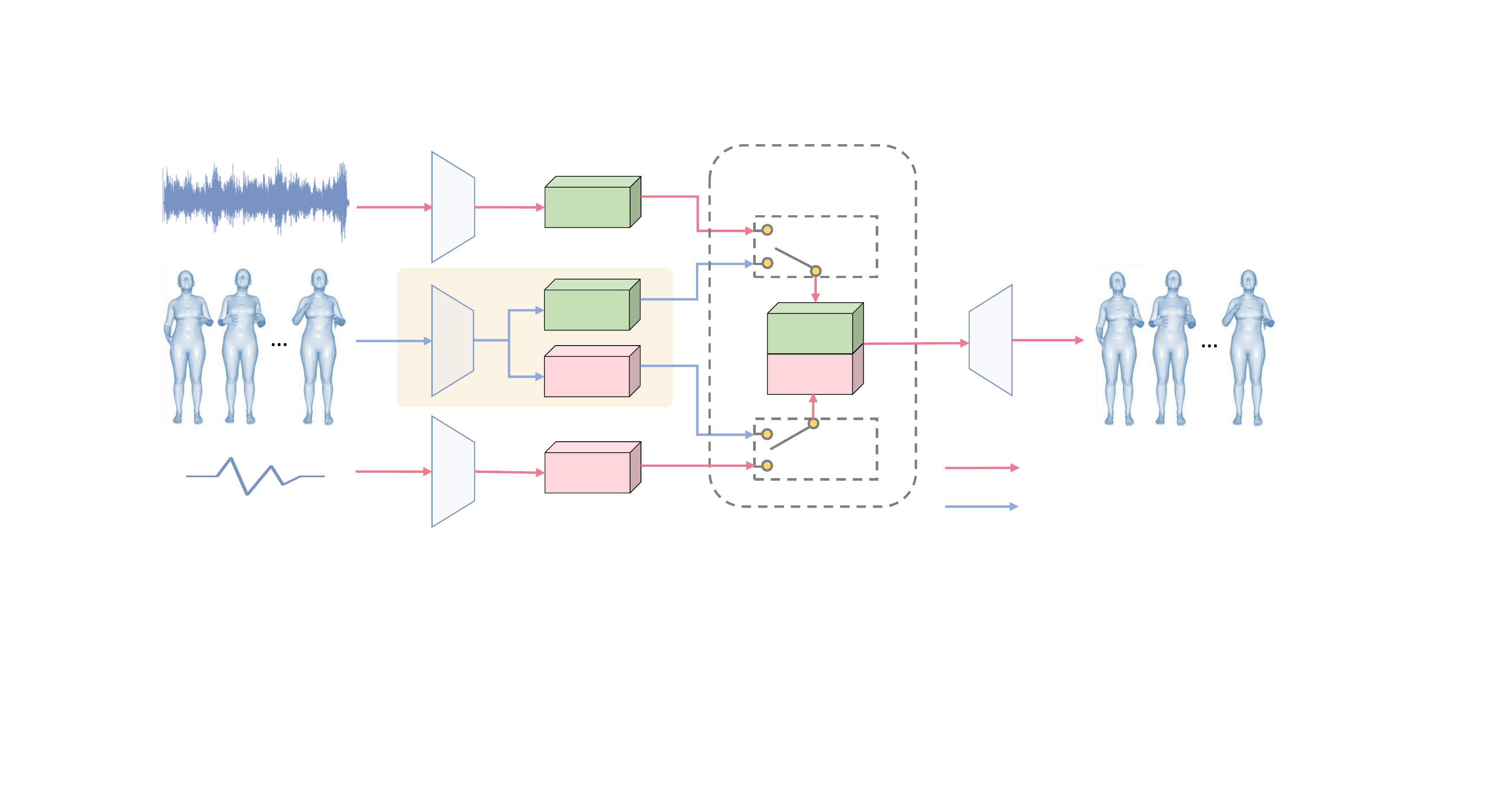}
\put(2.5,1){\small Random Sampling}
\put(25.5,4.5){\small $f_\text{R}$}
\put(25.5,16.1){\small $f_\text{M}$}
\put(25.5,28){\small $f_\text{A}$}
\put(38,4.5){\small $I_\text{R}$}
\put(38,13){\small $I_\text{M}$}
\put(38,19){\small $S_\text{M}$}
\put(38,28){\small $S_\text{A}$}
\put(58,13){\small $I$}
\put(58,16.6){\small $S$}
\put(73.6,16.6){\small $g$}
\put(50.5,30){\small Code Recombination}
\put(78,5){\small Training/Inference data flow}
\put(78,1.8){\small Training-only data flow}
\end{overpic}
\vspace{-6mm}
\caption{Our method explicitly models the audio-motion mapping by splitting the latent code into shared and motion-specific codes. The decoder generates different motions by recombining the shared and motion-specific codes extracted from different sources. The data flow in blue is only used at the training stage because we do not have motion data during inference.}
\label{fig:model}
\end{figure*}

%% file: img/fig_training.tex
\begin{figure}[!t]
\centering
\begin{overpic}[width=0.7\columnwidth,grid=false,trim=0.5cm 0.5cm 0.5cm 0.5cm,clip]{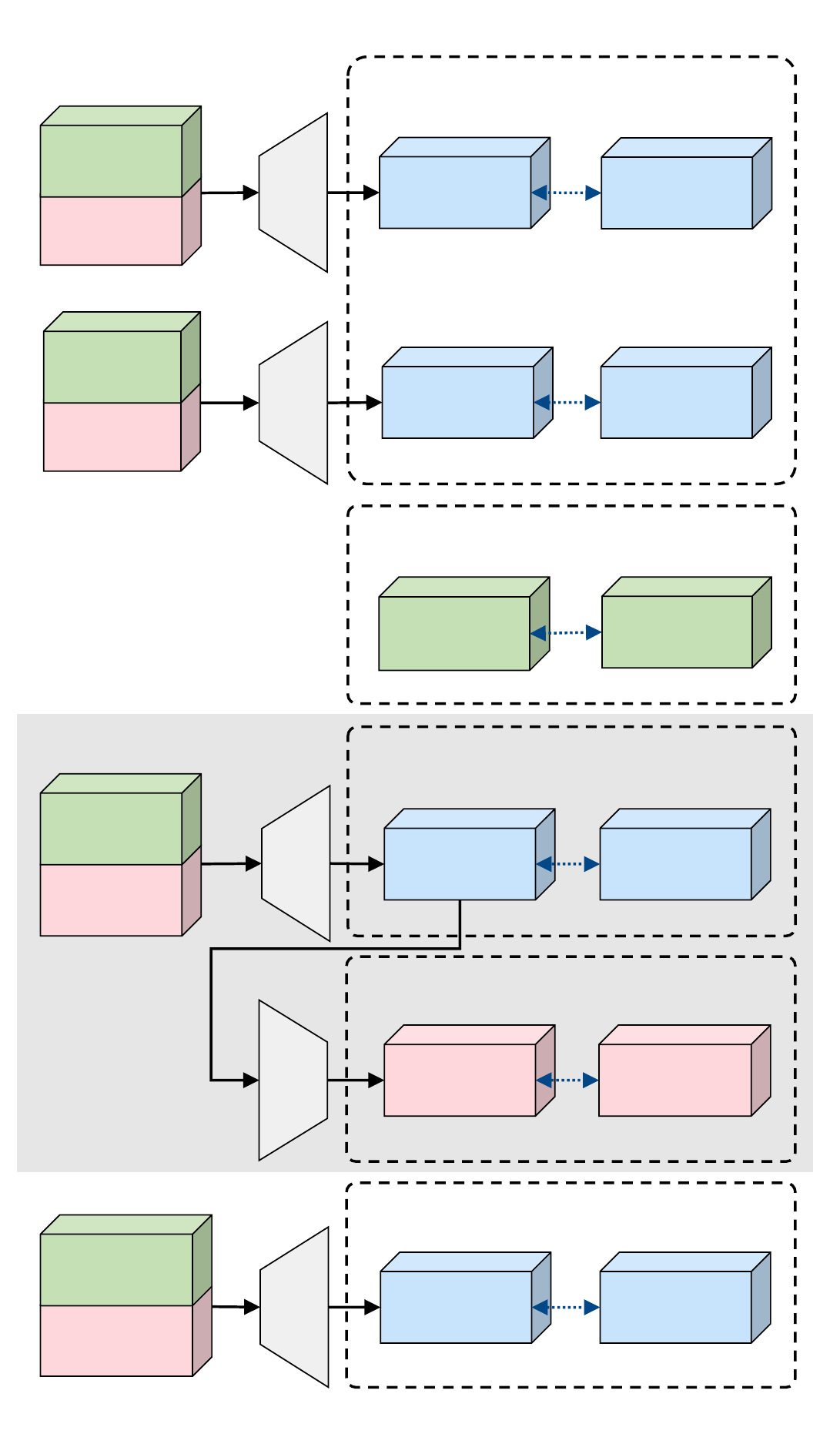}

\put(25,95){\small Motion Loss}
\put(4,90){\small $S_{\text{M}}$}
\put(4,85){\small $I_{\text{M}}$}
\put(18.5,88){\small $g$}
\put(29,88){\small $\hat{M}$}
\put(46,88){\small $M$}

\put(4,75){\small $S_{\text{A}}$}
\put(4,70){\small $I_{\text{M}}$}
\put(18.5,73){\small $g$}
\put(27,72){\small $\hat{M}_{S_{\text{A}}I_{\text{M}}}$}
\put(46,72){\small $M$}

\put(25,63){\small Alignment Constraint}
\put(29,55){\small $S_{\text{M}}$}
\put(46,55){\small $S_{\text{A}}$}

\put(25,46){\small Relaxed Motion Loss}
\put(4,41){\small $S_{\text{A}}$}
\put(4,36){\small $I_{\text{R}_1}$}
\put(18.5,39){\small $g$}
\put(46,39){\small $M$}
\put(27,39){\small $\hat{M}_{S_{\text{A}}I_{\text{R}_1}}$}

\put(25,29){\small Bicycle Constraint}
\put(17,23){\small $f_{\text{M}}$}
\put(29,23){\small $\hat{I}_{\text{R}_1}$}
\put(46,23){\small $I_{\text{R}_1}$}

\put(25,12){\small Diversity Loss}
\put(4,4){\small $I_{\text{R}_2}$}
\put(4,9){\small $S_{\text{A}}$}
\put(18.5,6){\small $g$}
\put(44,6){\small $\hat{M}_{S_{\text{A}}I_{\text{R}_1}}$}
\put(27,6){\small $\hat{M}_{S_{\text{A}}I_{\text{R}_2}}$}
\end{overpic}

\vspace{-4mm}
\caption{The training details of our model. 
Our model is trained with alignment constraint, motion reconstruction losses, relaxed motion loss, bicycle constraint, diversity loss, and KL divergence 
% \dknote{6 losses. KL is not in the figure.}. 
%
The alignment constraints and motion reconstruction loss help the model learn the audio-motion joint embedding.
The relaxed motion loss avoids the degeneration of the shared code.
The bicycle constraints and the diversity loss help reduce the mode-collapse problem and guide the model to generate multimodal motions.
The KL divergence is omitted in the figure for the sake of brevity.
}
\label{fig:training}
\end{figure}

%% file: tab_tex/sota.tex
\begin{table*}[!t]

\caption{
\textbf{Comparisons with previous SOTA methods} on Trinity dataset and S2G-Ellen dataset.
$\uparrow$ means the higher is better and $\downarrow$ means the lower is better. For methods supporting sampling, we run 20 tests and report their {\it average} score and the {\it best} score (in parentheses). Speech2Gesture (``S2G'' in the table) could not generate multimodality motions.
See the accompanied video in our supplementary.
}
\label{tab:sota_result}
\vspace{-4mm}
{\def\arraystretch{1} \tabcolsep=2.2em 
\begin{tabular}{ll|r@{\hspace{0.1cm}}lr@{\hspace{0.1cm}}lr@{\hspace{0.1cm}}lc}
\toprule[1pt]
Dataset & Method & \multicolumn{2}{c}{Pos. $L_1\downarrow$}  & \multicolumn{2}{c}{PCK $\uparrow$} & \multicolumn{2}{c}{Diverisity $\uparrow$ }& Multimodality $\uparrow$ \\
\midrule[1pt]
\multirow{5}{*}{\specialcell[c]{Trinity \\ (3D)}} & S2G w/o GAN \cite{speech2gesture} & 7.71 &  & 0.82 & & 5.99&  & - \\
& S2G \cite{speech2gesture} & 24.68 &  & 0.39 &  & 2.46 & &  - \\
& StyleGestures \cite{stylegestures} & 18.97 & (18.07) & 0.34 & (0.34) & 2.34 & (3.79) & \textbf{7.55} \\
& Audio2Gestures (Ours) & \textbf{7.84} & \textbf{(7.65)} & \textbf{0.82} & \textbf{(0.83)} & \textbf{6.32} & \textbf{(6.52)} & 4.11 \\
% & \lj{Ours w/ DCT} & & & & &  \\
\midrule[0.8pt]
\multirow{4}{*}{\specialcell[c]{S2G-Ellen \\ (2D)}} & S2G w/o GAN \cite{speech2gesture} & \textbf{0.74} & & \textbf{0.37} &  & 0.61 &  & - \\
& S2G \cite{speech2gesture} & 1.08 &  & 0.23 &  & \textbf{0.89} & & - \\
& Audio2Gestures (Ours) & 0.94 & (0.92) & 0.33 & (0.34) & 0.84 & (0.85) & 0.77 \\
% & \lj{Ours w/ DCT} & \\
% \midrule[0.8pt]
% \multirow{4}{*}{PhantomDance-100} & MINT~\cite{aist++} \\
% & DanceRevolution~\cite{dance_revolution} \\
% & Ours w/o DCT \\
% & Ours w/ DCT \\
% \midrule[0.8pt]
% \multirow{4}{*}{DanceRevolution} & MINT~\cite{aist++} \\
% & DanceRevolution~\cite{dance_revolution} \\
% & Ours w/o DCT \\
% & Ours w/ DCT\\
\bottomrule[1pt]
\end{tabular}}

% $\uparrow$ means the higher is better and $\downarrow$ means the lower is better. For methods supporting sampling, we run 20 tests and report their {\it average} score and the {\it best} score (in parentheses). Speech2Gesture (``S2G'' in the table) could not generate multimodality motions. 
% \dknote{how do you (experimentally) prove the effectiveness of the DCT module. add a subsection about it. E.g. test the spectral density of the joint trajectories?}
% \dknote{Other than DCT, any other (method-wise) improvements could be added?}
\end{table*}

%% file: tab_tex/ablation_components.tex
\begin{table*}[!t]

\caption{
\textbf{Ablation study of the proposed components} on the Trinity dataset.
Note that every line adds a new component compared to its previous line, so the last line is our full A2G model. 
For methods supporting sampling, we run 20 tests and report their {\it average} score and the {\it best} score (in parentheses). 
See Fig.~\ref{fig:user_study_abla} for user study results and the accompanied video in our supplementary.
}
\label{tab:ablation_components}
\vspace{-4mm}
{\def\arraystretch{1}\tabcolsep=0.2em
\begin{tabular}{l|
% r@{\hspace{0.1cm}}l
r@{\hspace{0.1cm}}l
r@{\hspace{0.1cm}}l
r@{\hspace{0.1cm}}l
r@{\hspace{0.1cm}}l|
r@{\hspace{0.1cm}}l
r@{\hspace{0.1cm}}l|
r@{\hspace{0.1cm}}l
r@{\hspace{0.1cm}}l|
r@{\hspace{0.1cm}}l
c}
%   \begin{tabular}{l|rlrlrlrlc}
\toprule[1pt]
Method & \multicolumn{2}{c}{Pos. $L_1$ $\downarrow$}
% & \multicolumn{2}{c}{Pos. $L_1$ (Norm.) $\downarrow$}
& \multicolumn{2}{c}{Speed $\downarrow$}
& \multicolumn{2}{c}{Acc. $\downarrow$}
& \multicolumn{2}{c|}{ PCK $\uparrow$ }
& \multicolumn{2}{c}{STFT $\downarrow$} 
& \multicolumn{2}{c|}{SSIM$\uparrow$} 
& \multicolumn{2}{c}{LPIPS $\downarrow$}
& \multicolumn{2}{c|}{FID $\downarrow$}
& \multicolumn{2}{c}{Diversity $\uparrow$}
& Multimodality $\uparrow$ \\
\midrule[1pt]
baseline & \textbf{7.64}  & (7.63) & \textbf{0.60} & (0.60) & \textbf{0.20} & (0.16) & \textbf{0.82} & (0.82)  & 1.74 & (1.74)  & \textbf{0.8765} & (0.8766)  & 46.3 & (46.3)  & 2.94 & (2.93)  & 5.35 & (5.37) & 0.41 \\
% \hspace{0.1cm}+ split & 12.1 & (11.9) & 1.48 & (1.46) & 2.36 & (2.03) & 0.57 & (0.58) & 2.27 & (2.27) & 0.7699 & (0.7714) & 85.9 & (85.1) & 10.7 & (10.6) & 6.41 & (6.46) & 3.37 \\
\hspace{0.1cm}+ split & 8.30 & (7.86) & 0.68 & (0.63) & 0.24& (0.18) & 0.79 & (0.82) & 1.12 & (1.11) & 0.8567 & (0.8665) & 52.1 & (48.6) & 2.94 & (2.16) & 5.03 & (6.09) & 5.95 \\
\hspace{0.1cm}+ mapping net & 7.73 & (7.45) & 0.67 & (0.66) &0.24&(0.21) & 0.81 & (0.82) & \textbf{1.11} & (1.10) & 0.8551 & (0.8563) & 47.8 & (47.1) & 2.19 & (1.86)  &  6.13 & (6.53) & 3.59 \\
\hspace{0.1cm}+ bicycle constraint & 7.69 & (7.36) & 0.64 & (0.62) & 0.24 & (0.20) & 0.82 & (0.84) & 1.14 & (1.14) & 0.8659 & (0.8700) & \textbf{46.3} & \textbf{(43.7)} & \textbf{2.16} & (1.70) & 6.35 & (6.67) & 3.99 \\
\hspace{0.1cm}+ diversity loss &  7.86 & (7.55) &  0.67 & (0.66) & 0.24 & (0.22) & 0.81 & (0.83)  & 1.13 & (1.12) & 0.8594 & (0.8619)  & 48.6 & (45.4) & 2.22 & (1.65) & \textbf{6.58} & (6.90) & \textbf{4.10} \\
\bottomrule[1pt]
\end{tabular}}
% \vspace{1mm}
% \begin{tablenotes}
% \small
% Note that every line adds a new component compared to its previous line. For methods supporting sampling, we run 20 tests and report their {\it average} score and the {\it best} score (in parentheses). 
% \dknote{may need a full ablation study on PhantomDance-100 too.}
% \end{tablenotes}
\end{table*}

%% file: tab_tex/ablation_relax_motion_loss.tex
\begin{table*}[!t]

\caption{
\textbf{Influence of using different threshold $\rho$ (in cm) in the relaxed motion loss for training.}
The best result of every metric is in bold and the second best result is underlined.
MM represents multimodality.
For easier comparison, we include their corresponding curves at the bottom, where the horizontal axis represents $\rho$ from 0 to 5.
We used $\rho=0.2$ by default.}
\label{tab:relaxed_motion_loss}
\vspace{-4mm}
{\def\arraystretch{1}\tabcolsep=0.33em
\begin{tabular}{l|
% r@{\hspace{0.1cm}}l
r@{\hspace{0.1cm}}l
r@{\hspace{0.1cm}}l
r@{\hspace{0.1cm}}l
r@{\hspace{0.1cm}}l|
r@{\hspace{0.1cm}}l
r@{\hspace{0.1cm}}l|
r@{\hspace{0.1cm}}l
r@{\hspace{0.1cm}}l|
r@{\hspace{0.1cm}}l
c}
\toprule[1pt]
$\rho$
&\multicolumn{2}{c}{Pos. $L_1$ $\downarrow$}
% &\multicolumn{2}{c}{Pos. $L_1$ (Norm.) $\downarrow$}
&\multicolumn{2}{c}{Speed $L_1$ $\downarrow$}
&\multicolumn{2}{c}{Acc. $L_1$ $\downarrow$}
&\multicolumn{2}{c|}{PCK $\uparrow$}
&\multicolumn{2}{c}{STFT $\downarrow$}
&\multicolumn{2}{c|}{SSIM $\uparrow$}
&\multicolumn{2}{c}{LPIPS $\downarrow$}
&\multicolumn{2}{c|}{FID $\downarrow$}
&\multicolumn{2}{c}{Diversity $\uparrow$}
& MM $\uparrow$\\
\midrule[1pt] 
0 & \textbf{7.4421} & (7.2117) & \textbf{0.6345} & (0.6163) & 0.2315 &(0.2059)& \textbf{0.8403} & (0.8541) & \uline{1.11} & (1.10) & \uline{0.8679} & (0.8720) & \textbf{44.6} & (42.8) & 2.15 & (1.80) & 6.15 & (6.45) & 3.51 \\
0.1 & \uline{7.6120} & (7.3529) & \uline{0.6295} & (0.6235)  & \textbf{0.2259} &(0.1919) & 0.8208 & (0.8329) & 1.11 & (1.11) & \textbf{0.8682} & (0.8704) & 46.3 & (45.0) & 2.25 & (2.09) & 6.20 & (6.32) & 2.70 \\
0.2 & 7.8613 & (7.5515) &  0.6678 & (0.6608) & 0.2432 & (0.2195) & 0.8130 & (0.8286) & 1.13 & (1.12) & 0.8594 & (0.8619)  & 48.6 & (45.4) & 2.22 & (1.65) & \uline{6.58} & (6.90) & \uline{4.10} \\
0.5 & 7.5915 & (7.4267) & 0.6509 & (0.6378) & \uline{0.2340} & (0.2018) & 0.8264 & (0.8388) & \textbf{1.09} & (1.08) & 0.8605 &(0.8677) & 46.2 & (45.1) & \textbf{2.06} & (1.66) & 6.51 & (6.74) & 3.33  \\
1 & 7.8480 & (7.5192) & 0.6688 & (0.6494) & 0.2507 & (0.2259) & 0.8133 & (0.8281) & 1.15 & (1.14) & 0.8596 & (0.8640) & 47.2 & (44.7) & \uline{2.07} & (1.66) & 6.43 & (6.60) & 4.02 \\
2 & 7.8575 & (7.4219) & 0.6561 & (0.6402)  & 0.2428 & (0.2062) & \uline{0.8169} & (0.8370) & 1.14 & (1.14) & 0.8625 & (0.8678) & 46.8 & (44.0) & 2.16 & (1.59) & 6.23 & (6.66) & \textbf{4.24} \\
5 & 7.7102 & (7.5064)& 0.6631 & (0.6353) & 0.2523 & (0.2063) & 0.8217 & (0.8352) & 1.16 & (1.14) & 0.8569 & (0.8582) & \uline{45.6} & (44.8) & 2.14 & (1.80) & \textbf{6.58} & (7.16) & 3.92 \\
\midrule
&\multicolumn{2}{l}{\includegraphics[width=1cm]{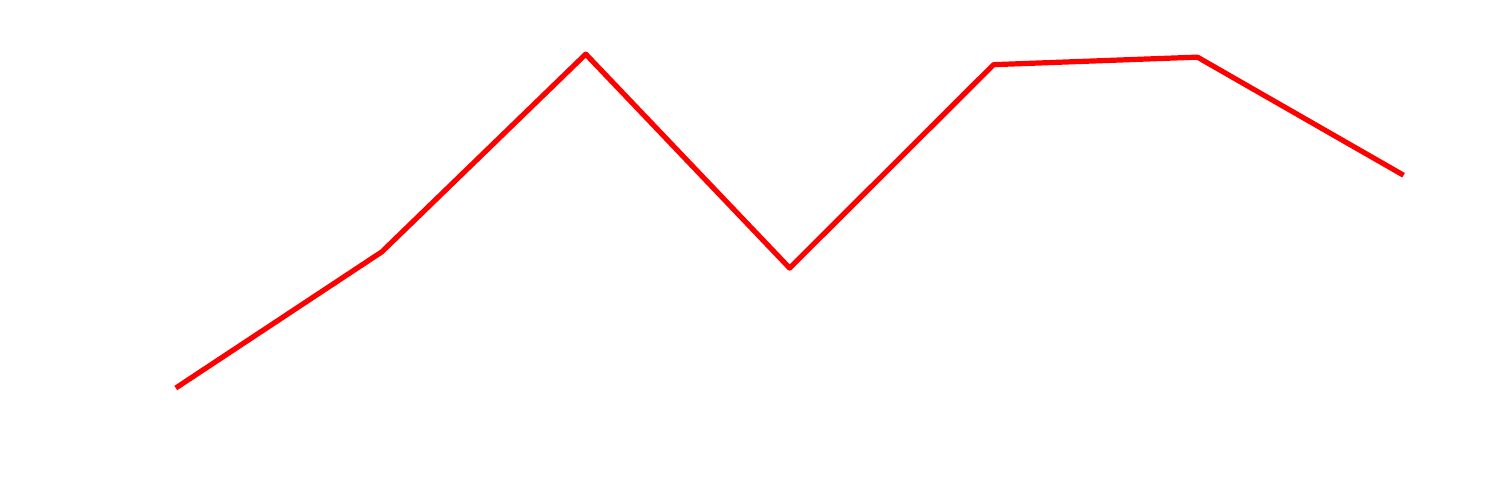}}&\multicolumn{2}{l}{\includegraphics[width=1cm]{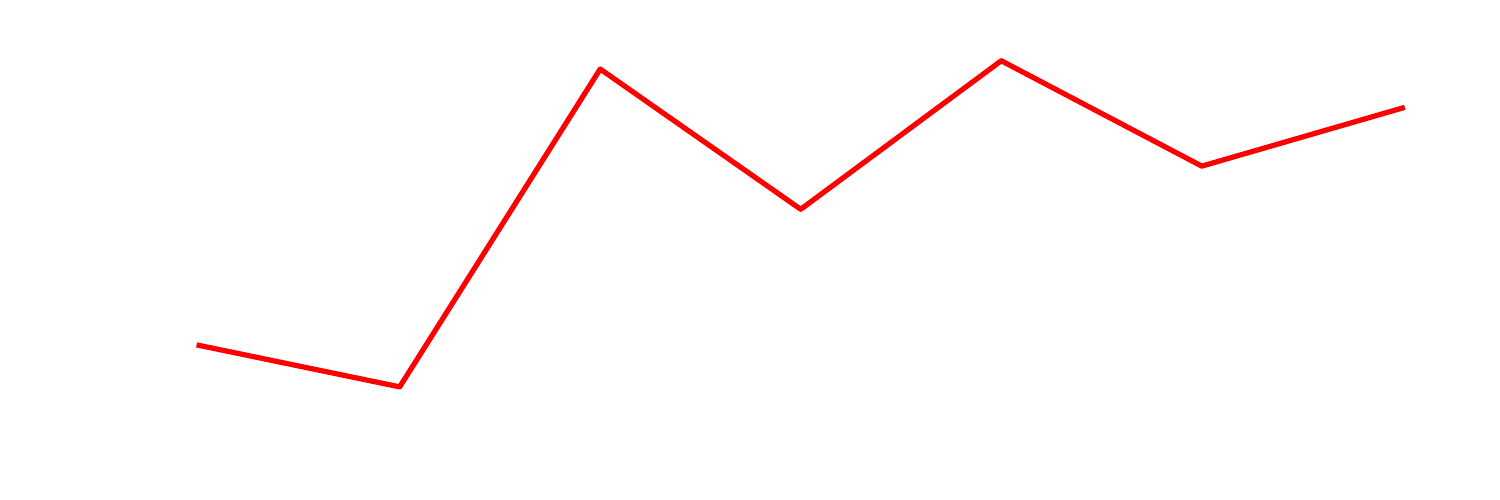}}&\multicolumn{2}{l}{\includegraphics[width=1cm]{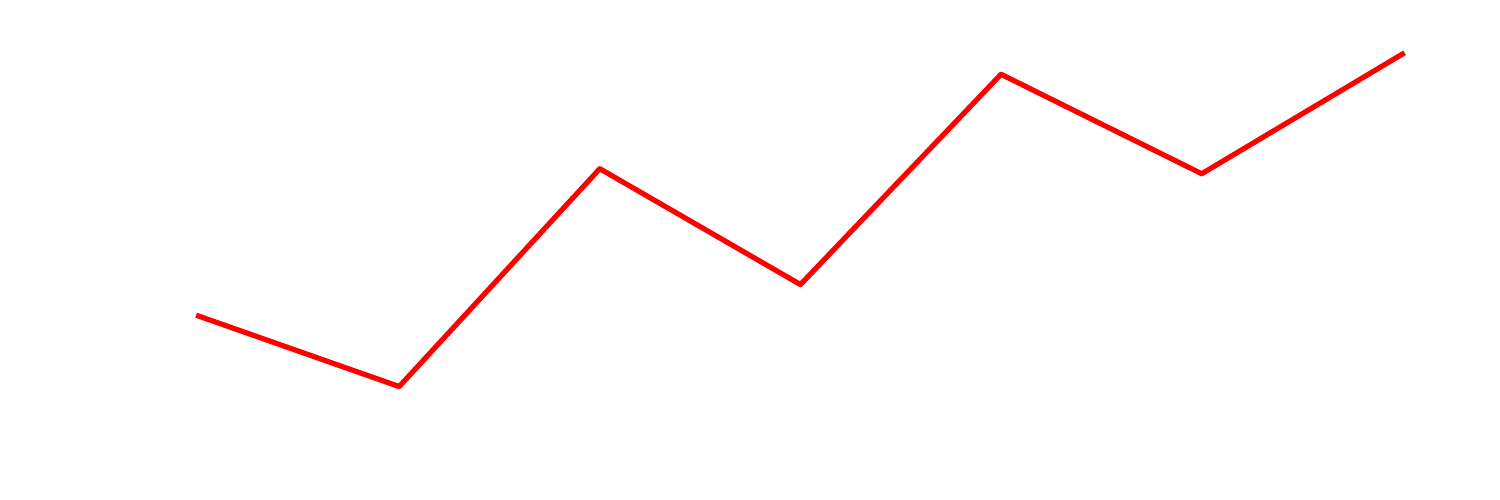}}&\multicolumn{2}{l|}{\includegraphics[width=1cm]{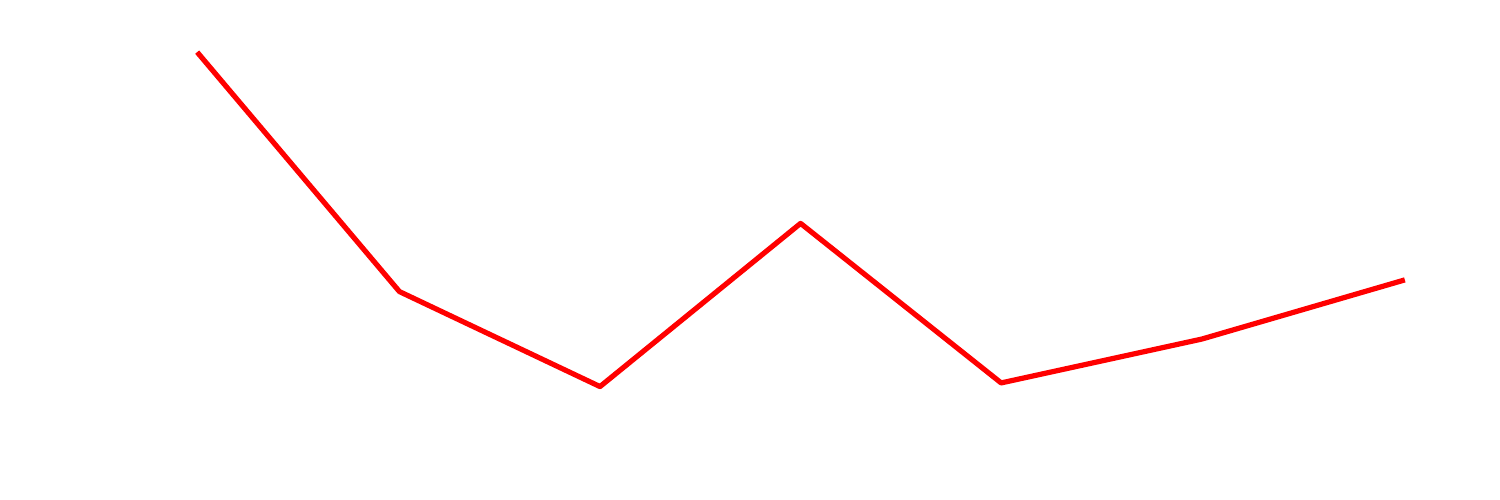}}&\multicolumn{2}{l}{\includegraphics[width=1cm]{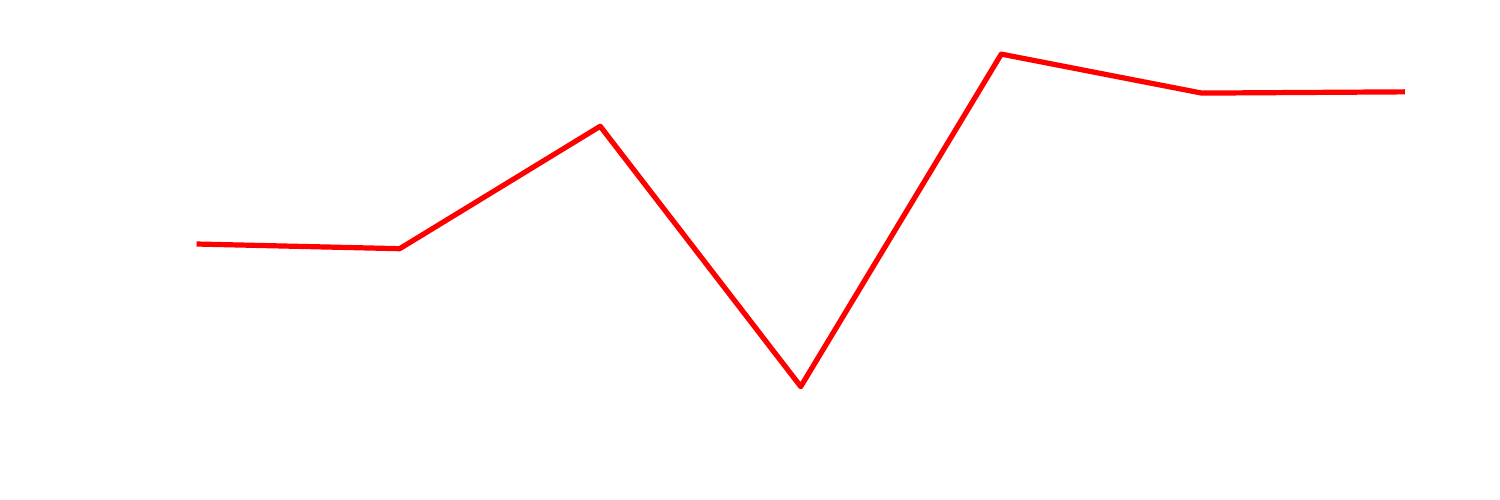}}&\multicolumn{2}{l|}{\includegraphics[width=1cm]{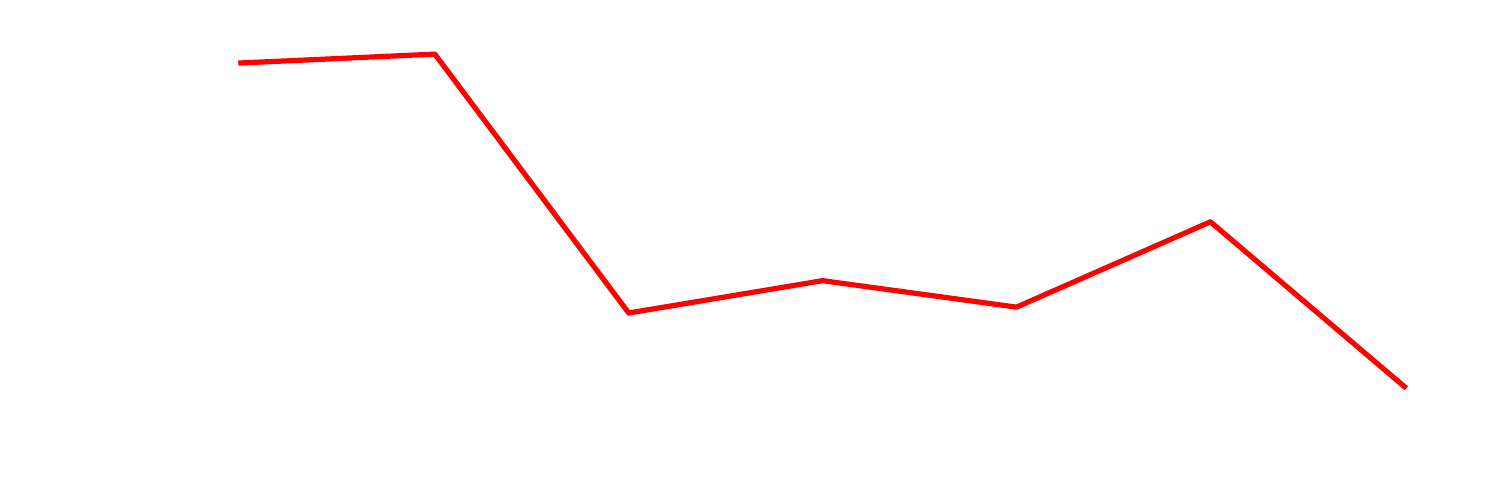}}&\multicolumn{2}{l}{\includegraphics[width=1cm]{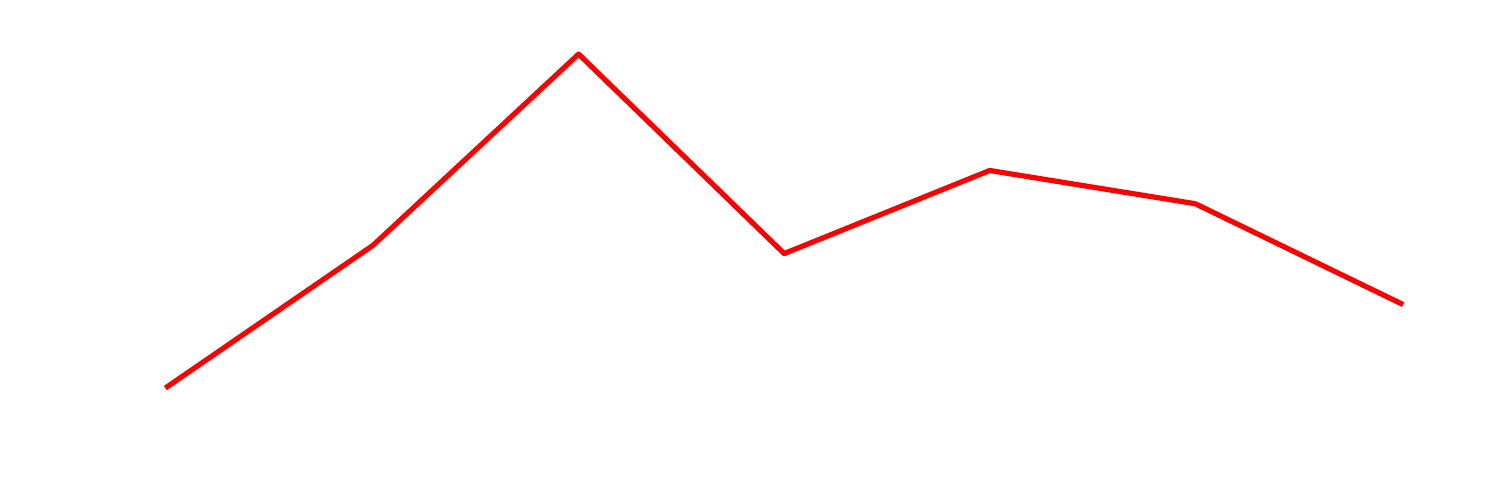}}&\multicolumn{2}{l}{\includegraphics[width=1cm]{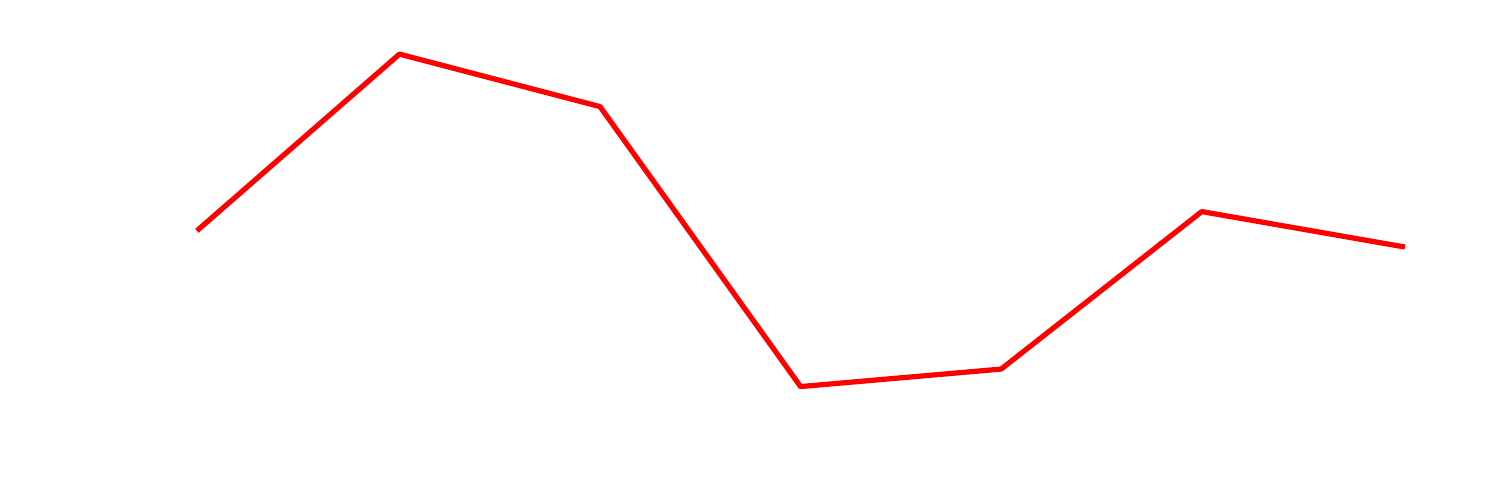}}&\multicolumn{2}{l}{\includegraphics[width=1cm]{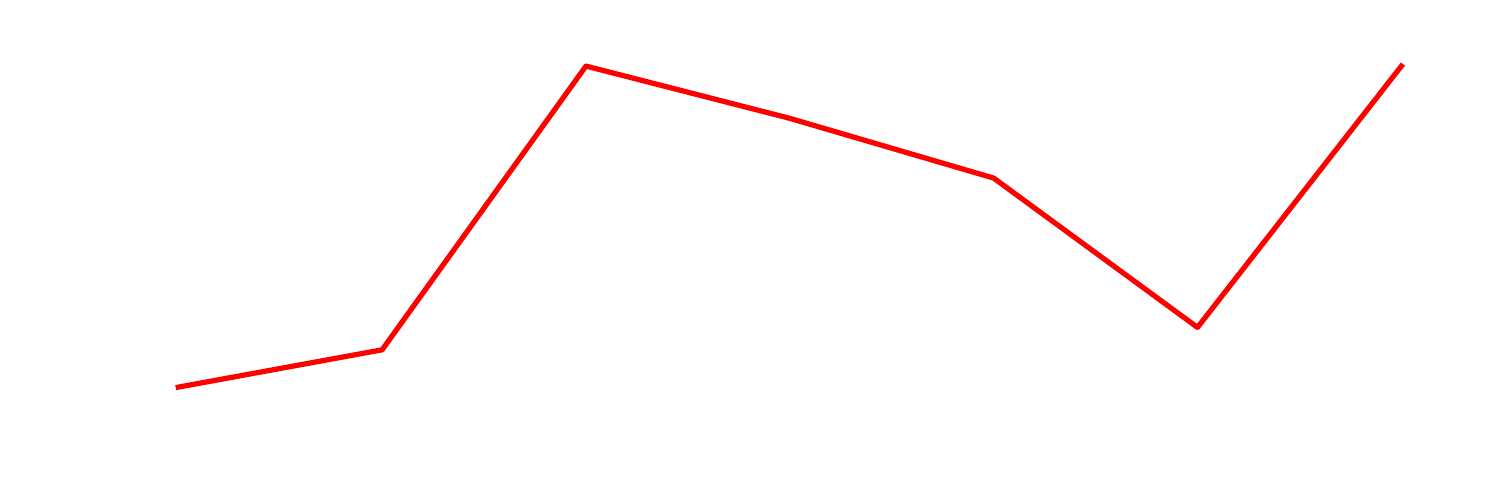}}&\includegraphics[width=1cm]{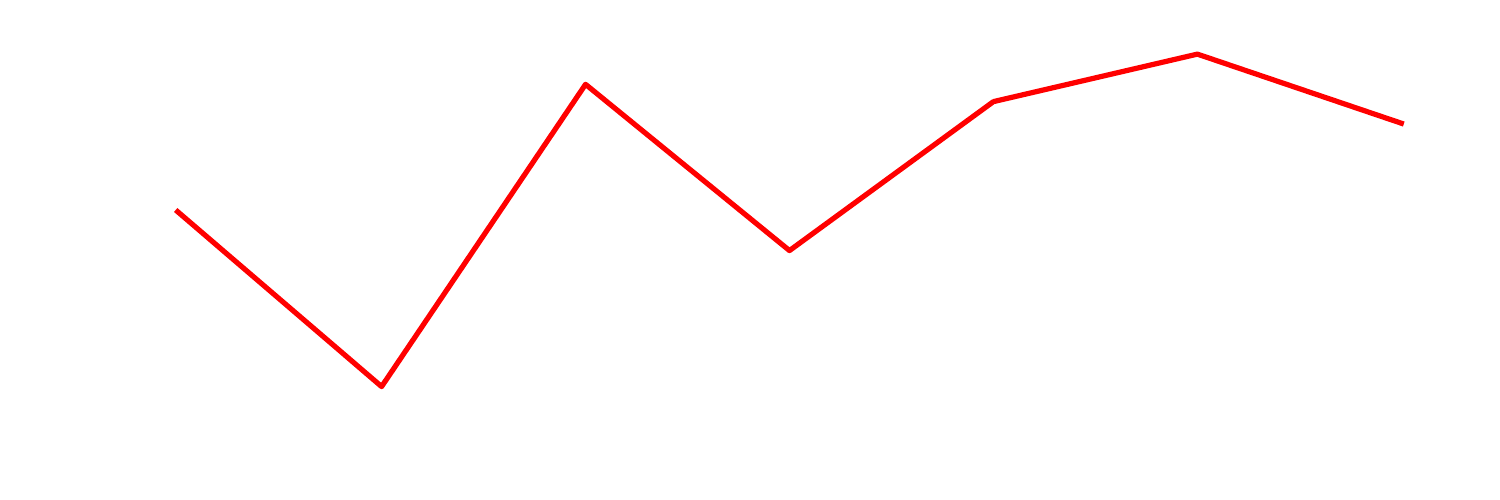}\\
\bottomrule[1pt]
\end{tabular}
}
\end{table*}

%% file: tab_tex/ablation_loss.tex
\begin{table*}[t]

\caption{
\textbf{Effect of introducing complementary motion losses.}
``+ loss\_x'' means including loss\_x to ``A2G + loss\_x''.
For methods supporting sampling, we run 20 tests and report their {\it average} score and the {\it best} score (in parentheses). Entries with improved performance are highlighted in \textcolor{blue}{blue}. 
MM represents multimodality.
See the accompanied video in our supplementary.
}
\label{tab:ablation_loss}
\vspace{-4mm}
{\def\arraystretch{1}\tabcolsep=0.2em
\begin{tabular}{
l|
r@{\hspace{0.1cm}}l
r@{\hspace{0.1cm}}l
r@{\hspace{0.1cm}}l
r@{\hspace{0.1cm}}l|  
r@{\hspace{0.1cm}}l
r@{\hspace{0.1cm}}l|
r@{\hspace{0.1cm}}l
r@{\hspace{0.1cm}}l|
r@{\hspace{0.1cm}}l
c
}
\toprule[1pt]
Basenet & 
\multicolumn{2}{c}{Pos. $L_1$ $\downarrow$} & 
% \multicolumn{2}{c}{Pos. $L_1$ (Norm.) $\downarrow$} & 
\multicolumn{2}{c}{Speed $L_1$ $\downarrow$} & 
\multicolumn{2}{c}{Acc. $L_1$ $\downarrow$} &
\multicolumn{2}{c|}{PCK $\uparrow$ } & 
\multicolumn{2}{c}{STFT $\downarrow$} & 
\multicolumn{2}{c|}{SSIM$\uparrow$} & 
\multicolumn{2}{c}{LPIPS $\downarrow$ } & 
\multicolumn{2}{c|}{FID $\downarrow$} &
\multicolumn{2}{c}{Diversity $\uparrow$} & 
MM $\uparrow$ \\
\midrule[1pt]
A2G & 7.8613 & (7.5515) &  0.6678 & (0.6608) & 0.2432 & (0.2195) & 0.8130 & (0.8286)  & 1.13 & (1.12) & 0.8594 & (0.8619)  & 48.6 & (45.4) & 2.22 & (1.65)& 6.58 & (6.90) & 4.10 \\
+ STFT & 7.9406 & (7.6353) & \cellcolor{blue!25}0.6565 & (0.6397) & \cellcolor{blue!25}0.2255 & (0.1895) & \cellcolor{blue!25}0.8134 & (0.8369) 
& \cellcolor{blue!25}1.02 & (1.00) & \cellcolor{blue!25}0.8621 & (0.8668) 
& 49.8 & (46.6) & 2.88 &(2.40) 
& 5.06 & (5.82) 
& \cellcolor{blue!25}4.88 \\
+ SSIM & \cellcolor{blue!25}7.8623 & (7.5101)& \cellcolor{blue!25}0.6384 & (0.6239) & \cellcolor{blue!25}0.2211 & (0.1803) & \cellcolor{blue!25}0.8167 & (0.8326) 
& \cellcolor{blue!25}1.10 & (1.09) & \cellcolor{blue!25}0.8647 & (0.8694) 
& 49.9 & (45.2)  & 2.98 & (2.37) 
& 5.19 & (6.09) 
& \cellcolor{blue!25}4.83 \\
+ LPIPS & 8.4157 & (7.9884) & 0.6680 & (0.6147)& 0.2521 & (0.1836) & 0.7895 &(0.8132) & 1.17 & (1.16) & 0.8551 & (0.8683)  &  \cellcolor{blue!25}48.3 & (43.5) & \cellcolor{blue!25} 1.90 & (0.95) & 6.04 & (6.86) & \cellcolor{blue!25}5.74  \\
\midrule[1pt]
GT+ $\mathcal{N}$(0,1)& 1.1617 & & 1.2352 & & 2.1369 & & 1.00 & & 2.78 & & 0.8449& & 0.51 & & 0.0004 & & 10.1 & & - \\ 
GT+ $\mathcal{N}$(0,5)& 4.4406 & & 6.0598 & & 10.503 & & 0.99 & & 4.06 & & 0.2749 & & 12.2 & & 0.2001 & & 11.7 & & - \\ 
\bottomrule[1pt]
\end{tabular}
}
% \lj{
% % Note that the baseline model is only trained with the motion reconstruction loss and the latent code learning related losses.
% % Other variations additionally introduce the corresponding loss.
% ``+ loss\_x'' means including loss\_x to A2G.For methods supporting sampling, we run 20 tests and report their {\it average} score and the {\it best} score (in parentheses).
% \dknote{add velocity loss}
% \dknote{explain number in parentheses}
% }
\end{table*}

%% file: tab_tex/backbone.tex
\begin{table*}[!t]
\caption{
\textbf{Comparisons of using different backbone networks.}
Trans. is short for Transformer. See the accompanied video in our supplementary.
}
\label{tab:backbone_ablation_study}
% \dknote{not very important if the Dance experiments go well.}
\vspace{-4mm}
{\def\arraystretch{1}\tabcolsep=0.3em
\begin{tabular}{l|
r@{\hspace{0.1cm}}l
r@{\hspace{0.1cm}}l
r@{\hspace{0.1cm}}l
r@{\hspace{0.1cm}}l|
r@{\hspace{0.1cm}}l
r@{\hspace{0.1cm}}l|
r@{\hspace{0.1cm}}l
r@{\hspace{0.1cm}}l|
r@{\hspace{0.1cm}}l
c}
\toprule[1pt]
Basenet & 
\multicolumn{2}{c}{Pos. $L_1$ $\downarrow$} &    
\multicolumn{2}{c}{Speed $L_1$ $\downarrow$} & 
\multicolumn{2}{c}{Acc. $L_1$ $\downarrow$} & 
\multicolumn{2}{c|}{PCK $\uparrow$} & 
\multicolumn{2}{c}{STFT $\downarrow$} & 
\multicolumn{2}{c|}{SSIM$\uparrow$} & 
\multicolumn{2}{c}{LPIPS $\downarrow$} & 
\multicolumn{2}{c|}{FID $\downarrow$} &
\multicolumn{2}{c}{Diversity $\uparrow$} & Multimodality $\uparrow$ \\
\midrule[1pt]
baseline & 7.64  & (7.63) & 0.60 & (0.60) & 0.20 & (0.16) & 0.81 & (0.81)  & 1.74 & (1.74)  & 0.8765 & (0.8766)  & 46.3 & (46.3)  & 2.94 & (2.93)  & 5.35 & (5.37) & 0.41  \\
\midrule[0.5pt]
A2G &  7.86 & (7.55) &  0.67 & (0.66) & 0.24 & (0.22) & 0.81 & (0.83)  & \textbf{1.13} & (1.12) & 0.8594 & (0.8619)  & 48.6 & (45.4)  & 2.22 & (1.65) & \textbf{6.58} & (6.90) & 4.10\\
A2G (GRU) & 8.05 & (7.77)  & 0.64 & (0.64) & 0.23 & (0.22) &  0.80 & (0.82) & 1.28 & (1.27) & 0.8629 & (0.8653) & 47.6 & (45.9) & 1.95 & (1.83)& 6.44 & (6.77) & \textbf{4.39} \\
A2G (Trans.) & \textbf{7.53} & (7.27) & \textbf{0.61} &(0.61) & \textbf{0.20} & (0.16) & \textbf{0.83} & (0.84)& 1.41 & (1.40) & \textbf{0.8753} & (0.8764) & 47.9 & (46.8) & 2.55 & (2.41) & 5.18 & (5.37) & 2.27  \\
\bottomrule[1pt]
\end{tabular}
}
\end{table*}

%% file: tab_tex/dct.tex
\begin{table*}[!t]
\caption{
\textbf{Results after various editing of the latent DCT feature.}
% Motion editing by editing the feature with DCT.
$S_A$ and $I_R$ both have 128 DCT components.
$S_A[N:] = 0$ means that we keep the lowest N components and manually set the remaining components to zero.
See the accompanied video in our supplementary.
}
\label{tab:feat_dct}
\vspace{-4mm}
{\def\arraystretch{1}\tabcolsep=0.25em
\begin{tabular}{l|
r@{\hspace{0.1cm}}l
r@{\hspace{0.1cm}}l
r@{\hspace{0.1cm}}l
r@{\hspace{0.1cm}}l|
r@{\hspace{0.1cm}}l
r@{\hspace{0.1cm}}l|
r@{\hspace{0.1cm}}l
r@{\hspace{0.1cm}}l|
r@{\hspace{0.1cm}}l
c}
\toprule[1pt]
Method &
\multicolumn{2}{c}{Pos. $L_1$ $\downarrow$} &
\multicolumn{2}{c}{Speed $L_1$ $\downarrow$} &
\multicolumn{2}{c}{Acc. $L_1$ $\downarrow$} &
\multicolumn{2}{c|}{PCK $\uparrow$ }&
\multicolumn{2}{c}{STFT $\downarrow$}&
\multicolumn{2}{c|}{SSIM$\uparrow$}& 
\multicolumn{2}{c}{LPIPS $\downarrow$}& 
\multicolumn{2}{c|}{FID $\downarrow$}& 
\multicolumn{2}{c}{ Diversity $\uparrow$} &
Multimodality $\uparrow$
\\
\midrule
%     S2G~\cite{speech2gesture} & 7.71 &  & 0.82 & \\
A2G~\cite{audio2gestures} & 7.86 & (7.55) &  0.67 & (0.66) & 0.24 & (0.22) & 0.81 & (0.83)  & 1.13 & (1.12) & 0.8594 & (0.8619)  & 48.6 & (45.4)  & 2.22 & (1.65) & 6.58 & (6.90) & 4.10 \\
A2G w/ DCT & 8.22 & (7.99) & 0.68& (0.67) & 0.25 & (0.22) & 0.80 & (0.81) & 1.17 & (1.16) & 0.8558 & (0.8579) & 50.4 & (47.3) & 1.58 & (1.51) & 8.01 & (8.20) & 4.27 \\
\midrule
$S_A[100:]=0$ & 8.23 & (7.98) & 0.68 & (0.67) & 0.25 & (0.22) & 0.80 & (0.81) & 1.17 & (1.16) & 0.8557 & (0.8579) & 51.1 & (49.6) & 1.61 & (1.55) & 8.00 & (8.20) & 4.29 \\
$S_A[50:]=0$ & 8.24 & (7.99) & 0.68 & (0.67) & 0.25 & (0.22) & 0.79 & (0.81) & 1.15 & (1.15) & 0.8555 & (0.8577) & 51.6 & (49.8) & 1.58 & (1.53) & 8.02 & (8.22)  & 4.28\\
$S_A[10:]=0$ & 8.95 & (8.68)& 0.70 & (0.69) & 0.23 & (0.20) & 0.77 & (0.78) & 1.12 & (1.11) & 0.8445 & (0.8464) & 51.4 & (49.6) & 2.23 & (2.11) & 8.46 & (8.70) & 6.15 \\
\midrule
$I_R[10:]=0$ & 8.24 & (8.01) & 0.68 & (0.67) & 0.25 & (0.22) & 0.79 & (0.81) & 1.16 & (1.16) & 0.8555 & (0.8576) & 51.4 & (49.6) & 1.61 & (1.56) & 8.01 & (8.22) &4.27  \\
$I_R[1:]=0$ & 8.24 & (8.01) & 0.68 & (0.67) & 0.25 & (0.22) & 0.79 & (0.82) & 1.16 & (1.16) & 0.8556 & (0.8577) & 51.4 & (49.9) & 1.62 & (1.57)  & 8.02 & (8.24) & 4.25 \\
$I_R=0$ & 12.1 & (12.1) & 0.65 & (0.65) & 0.24 & (0.24) & 0.62 & (0.62) & 1.18 & (1.18) & 0.8432 & (0.8432) & 127.3 & (127.3) & 21.9 & (21.9) & 5.80 & (5.80) & - \\
\bottomrule[1pt]
\end{tabular}
}
% \lj{Note that $S_A$ and $I_R$ have 128 DCT components. $S_A[100:] = 0$ means that we only keep the lowest 100 components and manually set the highest 28 components to zero. }
\end{table*}